\documentclass[sn-mathphys,Numbered]{sn-jnl}

\usepackage{graphicx}%
\usepackage{multirow}%
\usepackage{amsmath,amssymb,amsfonts}%
\usepackage{amsthm}%
\usepackage{mathrsfs}%
\usepackage[title]{appendix}%
\usepackage{xcolor}%
\usepackage{textcomp}%
\usepackage{manyfoot}%
\usepackage{booktabs}%
\usepackage{algorithm}%
\usepackage{algorithmicx}%
\usepackage{algpseudocode}%
\usepackage{listings}%
\usepackage{array}
\usepackage{tabularx}
\usepackage{caption}

\theoremstyle{thmstyleone}%
\theoremstyle{thmstyletwo}%

\theoremstyle{thmstylethree}%

\begin{document}

\title[Article Title]{AraPoemBERT: A Pretrained Language Model for Arabic Poetry Analysis}


\author*[1]{\fnm{Faisal} \sur{Qarah}}\email{fqarah@taibahu.edu.sa}



\affil*[1]{\orgdiv{Department of Computer Science, College of Computer Science and Engineering}, \orgname{Taibah University}, \orgaddress{\city{Medina}, \postcode{42353}, \country{Saudi Arabia}}}




\abstract{Arabic poetry, with its rich linguistic features and profound cultural significance, presents a unique challenge to the Natural Language Processing (NLP) field. The complexity of its structure and context necessitates advanced computational models for accurate analysis. In this paper, we introduce AraPoemBERT, an Arabic language model pretrained exclusively on Arabic poetry text. To demonstrate the effectiveness of the proposed model, we compared AraPoemBERT with 5 different Arabic language models on various NLP tasks related to Arabic poetry. The new model outperformed all other models and achieved state-of-the-art results in most of the downstream tasks. AraPoemBERT achieved unprecedented accuracy in two out of three novel tasks: poet's gender classification (99.34\% accuracy), and poetry sub-meter classification (97.79\% accuracy). In addition, the model achieved an accuracy score in poems' rhyme classification (97.73\% accuracy) which is almost equivalent to the best score reported in this study. Moreover, the proposed model significantly outperformed previous work and other comparative models in the tasks of poems' sentiment analysis, achieving an accuracy of 78.95\%, and poetry meter classification (99.03\% accuracy), while significantly expanding the scope of these two problems. The dataset used in this study, contains more than 2.09 million verses collected from online sources, each associated with various attributes such as meter, sub-meter, poet, rhyme, and topic. The results demonstrate the effectiveness of the proposed model in understanding and analyzing Arabic poetry, achieving state-of-the-art results in several tasks and outperforming previous works and other language models included in the study. AraPoemBERT model is publicly available on \url{https://huggingface.co/faisalq}.}




\maketitle

\section{Introduction}\label{sec1}


The Arabic language is one of the world's most widely spoken languages. It has a rich history and its influence is seen across various domains including media, politics, history, and art. Arabic poetry, a prominent part of Arabic literature and culture, serves as a window to the norms, values, and historical events of the Arab world \cite{zwettler1978oral}. An Arabic poem, or \( qasida \), typically consists of one or more verses, each verse usually composed of two halves known as \( hemistiches \). All verses in a specific poem share the same rhyme and meter, creating a rhythmic pattern that adds to the beauty and depth of the poem. The rhyme, or \( qafia \), is a repeating pattern of sounds that occurs at the end of each verse, while the meter, or \( bahar \), dictates the rhythmic structure of the verse \cite{arberry1965arabic}. Arabic poetry spans a wide range of topics, encapsulating the poet's genuine emotions and thoughts. These topics can range from romance and longing to spiritual devotion, each offering a unique perspective and depth of emotion.

\begin{table}[t]
\parbox{.45\linewidth}{
\caption{List of classical meters in Arabic poetry.\label{tab1}}
\centering
\begin{tabular}{c}
\includegraphics[scale=0.25]{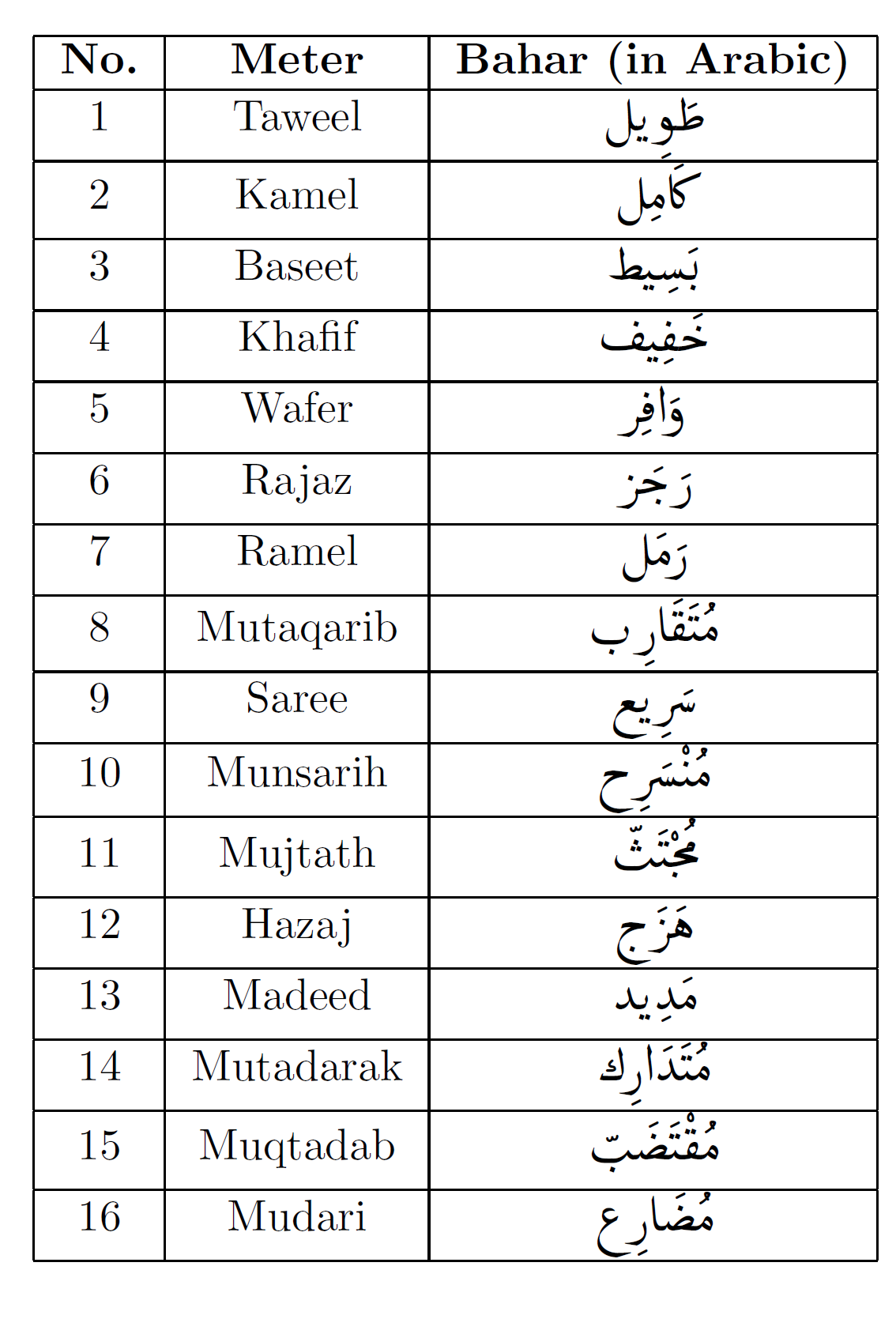}
\end{tabular}
} 
\hfill
\parbox{.45\linewidth}{ 
\caption{List of basic feet in Arabic poetry (Tafaeil).\label{tab2}}
\centering
\begin{tabular}{c}
\includegraphics[scale=0.25]{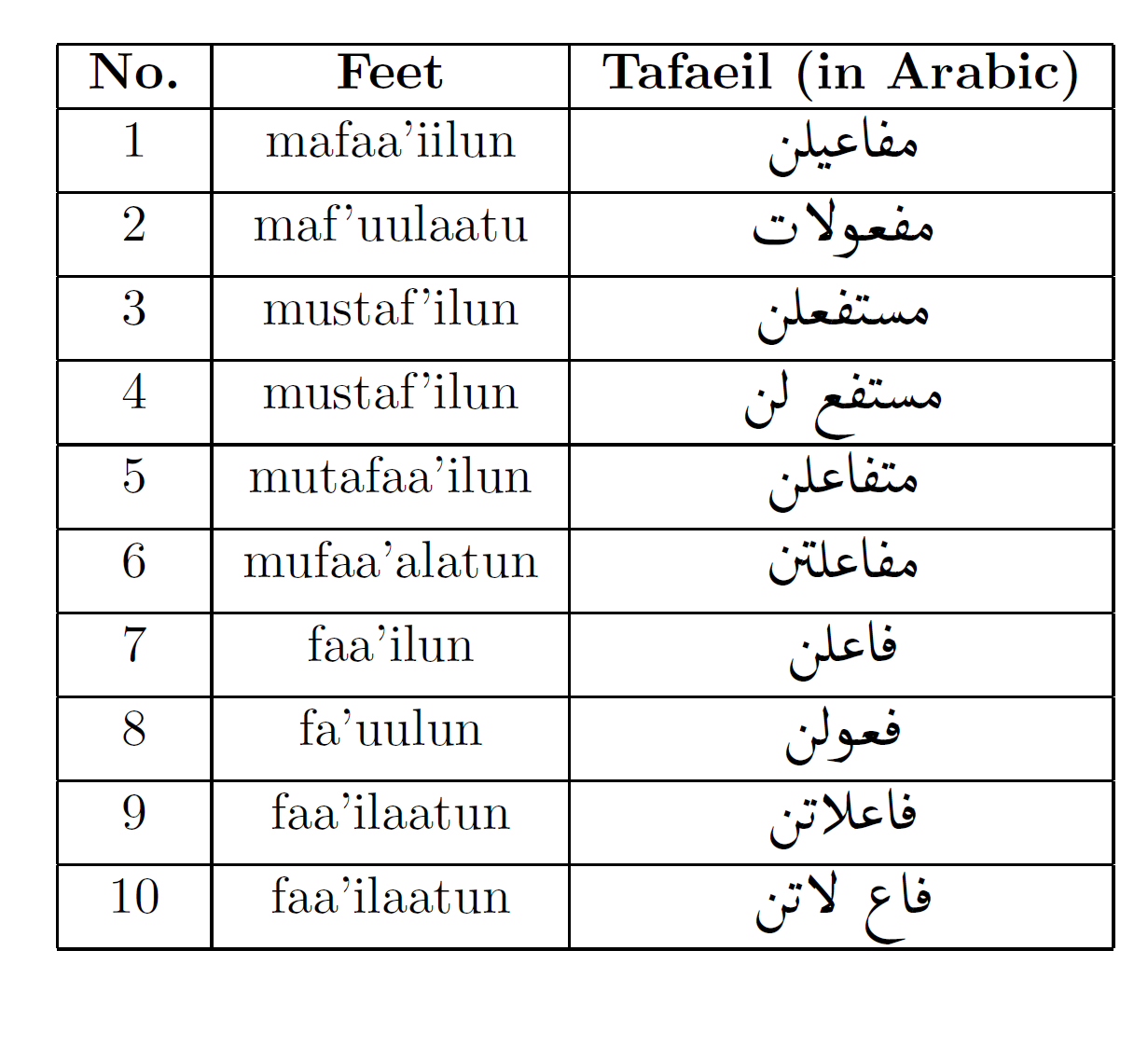}
\end{tabular}
}
\end{table}

Classical Arabic poetry adheres to a set of established meters, each with its unique rhythmic pattern. These meters are very important as they provide the poem with its rhythmic structure and flow \cite{scott2010pegs}. Table \ref{tab1} provides a list of all classical meters. The science of Arabic prosody, or \( 'Arud \), involves the use of \( 'Tafaeil \) or poetic feet. These are groups of ten expressions that scholars have agreed upon as the standard for weighing Arabic poetry. The ten feet in Arabic prosody, shown in Table \ref{tab2}, are composed of specific letters: Faa, Ain, Lam, Noon, Meem, Seen, Taa, and vowels. These feet correspond in weight to the letters of the measured words in the poem verse, matching vowels with vowels and consonants with consonants \cite{atiq1987elm}. However, modifications to these poetic feet can occur, altering their ideal image. These modifications can involve omitting, adding, or silencing parts of the feet. Scholars of 'Arud have differing views on these changes, with some approving and others disapproving \cite{ahmad2018causes}.

Typically, the meters used in Arabic poems are in their \( Complete \) form, meaning all the original Tafaeil of a specific meter are used, except for a few meters that must be in \( Majzuu \) (Fragmented) form. However, poets sometimes omit parts of the Tafaeil from the original meter, resulting in a variant of that meter. Scholars of 'Arud have identified seven different variants that could be derived from classical meters. Whereas some meters are always appear with a certain variant such as 'Complete', while others may come in more than one variant \cite{ahmad2018causes}. Table \ref{tab3} provides a brief description of the known meter variants. In this study, we will refer to the combination of meters and their existing variants as 'sub-meters', which include the name of the meter and its variant (e.g., Wafer-Majzuu, Rajaz-Complete, Kafif-Mahthuf, etc.).

\begin{table}
\caption{List of meter variants.~\label{tab3}}
\centering
\begin{tabular}{|c|p{5cm}|p{7cm}|}
\hline
\textbf{Variant Name} & \textbf{Description} & \textbf{Meters} \\
\hline
Complete & Has fulfilled all its original meter's feet & All meters can be in its complete meter form except: Madeed, Mudari, Hajaz, Muqtadab, and Mujtath \\
\hline
Majzuu & The last foot in each hemistich is omitted & Mandatory in: Madeed, Mudari, Hajaz, Muqtadab, and Mujtath. Also possible in all other meters except: Taweel, Munsarih, and Saree \\
\hline
Mashture & Dropping half the feet in both hemistiches & Can occur in Rajaz, Saree, and rarely in Baseet \\
\hline
Manhuk & Dropping two thirds the feet in both hemistiches & Can occur in Rajaz, Munsarih, and Mutadarak \\
\hline
Maktuu & Omitting the last syllable in the last foot and silencing what is before & Can only occur in Kamel meter \\
\hline
Ahuth & Only omitting the last half in the last foot & Can only occur in Kamel meter \\
\hline
Mukhala & A variation of Majzuu & Can only occur in Baseet meter \\
\hline
\end{tabular}
\end{table}

In contrast to classical meters, non-classical meters allow for more flexibility and diversity in constructing the poem. The usage of these patterns adds to the beauty of Arabic poetry, making it an enjoyable experience, either when reading silently or reciting out loud. Non-classical meters appeared chronologically after the classical meters and are often featured in \( Nabati \)  \( poetry \), which translates to folk poetry. The rise of these meters is closely associated with the prevalence of colloquial speech. The names and numbers of these meters vary among scholars, and their classification is often influenced by the era and region under consideration \cite{saad2010nabati}. However, identifying poems' meters manually poses a challenge. It necessitates an understanding of the language and its scientific study of 'Arud, as well as a keen ear for recognizing rhythm and sound patterns. This task becomes more demanding when dealing with non-classical rhythmic patterns (meters) that are more flexible and sometimes do not adhere to specific rules \cite{atiq1987elm}.

The goal of conducting a thorough analysis or solving different problems related to Arabic poetry has led to the development of various methods and techniques. One promising solution is the use of language models that can analyze and learn from text data. By pretraining a language model on a dataset of poems and fine-tuning it using verses text and their corresponding labels, we can create a system capable of accurately identifying the meter or rhyme of a given poem or a verse. This approach not only saves time and effort but also opens up new possibilities for analyzing and classifying Arabic poetry. In this paper, we present AraPoemBERT, a new BERT-based language model pretrained from scratch exclusively on Arabic poetry text. We provided a comprehensive evaluation of its performance in comparison with other Arabic language models on five different tasks related to Arabic poetry. We believe that AraPoemBERT has potential for the future of Arabic poetry analysis, serving as a valuable tool for scholars and researchers in linguistic, Arabic literature, and natural language processing (NLP) fields.

\noindent \\The main contributions of this paper can be summarized as follows:

\begin{itemize}

\item Presenting a new language model pretrained from scratch, dedicated solely to Arabic poetry.
\item Reporting state-of-the-art results in 4 out of 5 different NLP tasks related Arabic poetry using the proposed model compared to previous work and other prominent language models. 
\item We are the first to explore and report the results for 3 new tasks: poet's gender, poetry sub-meters, and poetry rhymes classification.
\item Compared to previous work, we have achieved significantly higher accuracy results in the tasks of classifying poetry meters, and poems sentiment analysis,  while expanding the scope of these problems.
\item The new dataset used in this study is the largest ever compiled, consisting of over 2.09M verses associated with various information.

\end{itemize}

The paper is organized as follows: Section 2 presents the related work. Section 3 introduces background about Transformers. In Section 4, we discuss the proposed model and the compiled dataset. The experimental procedure is presented in Section 5. The experimental results are discussed in Section 6. Finally, the conclusion is presented in Section 7.

\section{Related Work}\label{sec2}

\subsection{Arabic Poetry Analysis}
In recent years, the natural language processing (NLP) research related to Arabic poetry has focused mainly on two tasks: poems meters classification, and poems sentiment analysis using machine learning models.

Several researchers in the past have proposed rule-based algorithms aiming to classify poetry meters \cite{alabbas2014basrah}\cite{abuata2018rule}\cite{alnagdawi2013finding}. These approaches convert the input text into its Arudi form using regular expressions or Khashan's "numerical prosody" method, and subsequently they determine the meter of the target verse or poem. These systems heavily rely on diacritizing the input text and necessitate an understanding of the Arudi field to create effective rules to be used in these systems.
Berkani et al. \cite{berkani2020pattern} suggested a pattern recognition extraction and matching approach for poems meter detection. This method involves extracting a group of patterns from a target verse and comparing them to a set of labeled patterns. If the extracted pattern matches any of the labeled ones, the system can identify the meter of the input verse. The reported accuracy of this approach reached 99.3\% when tested on a dataset consisting of 2,711 verses. However, it should be noted that common poets' practices, such as text vocalization or minor imperfections in the poem, can potentially impact the system's accuracy.
Yousef et al. \cite{yousef2019learning} were the first to utilize machine learning models in addressing the problem of poem meter classification. They proposed a recurrent neural network (RNN) model to classify poems in Arabic and English languages. The model was trained and tested on a new dataset called APCD \cite{apcd} and consists of 1.83 million Arabic verses that were collected by the authors from online sources. The proposed model has achieved an impressive overall accuracy of 96.38\% when tested in classifying the 16 classical meters.
Similarly, Shaibani et al. \cite{alShaibani2020meter} proposed a novel approach utilizing five bidirectional gated recurrent unit (BiGRU) layers, and used character-based encoding for text representation. The researchers collected a set of poems comprising 55,440 verses categorized into 14 meters only, and achieved an overall accuracy of 94.32\%.

Abandah et al. \cite{abandah2022classifying} introduced a new machine learning model that contains four bidirectional long short-term memory (BiLSTM) layers. The proposed model was trained and tested on a revised version of APCD dataset called APCD2 that is composed of 1.6 million verses classified into 16 poem meters. Compared to \cite{yousef2019learning}, the new model is relatively smaller in size and is significantly better, achieving an accuracy of 97.27\%. Additionally, Abboushi et al. \cite{abboushi2023toward} proposed a poem generation model by fine-tuning AraGPT2 \cite{antoun2020aragpt2} using APCD2 dataset. When generating a poem text, the proposed model adheres to a specific meter and rhyme that can be detected from a given verse prompt. However, the overall accuracy of the model in classifying and detecting poetry meters is significantly lower than previous work, since the accuracy scores of classifying the meters that are underrepresented in the dataset are between 0\% and 17\%.

Similar to the attempts at classifying poem meters, numerous methods have been suggested in literature towards poems sentiment analysis. Mohammad \cite{mohammad2009naive} presented a Naive Bayes approach for classifying poems into seven different categories Hekmah (Wisdom), Retha (Elegy poems), Ghazal (Spinning poems), Madeh (Praise), Heja (Satire), Wasef (Description poems), Fakher (self-glorification), and Naseeb (contentment) using 20 Arabic poems with six verses each, and had achieved an accuracy of 55\%.
Alsharif et al. \cite{alsharif2013emotion} classified Arabic poems into four classes: Retha (Elegy poems), Ghazal (Spinning poems), Fakhr (self-glorification), and Heja (Satire) using Naive Bayes and support vector machine (SVM) models. They used a dataset composed of 1231 poems comprising 20041 verses, and they have achieved an F1-score of 0.66 as the highest result reported in their work.
Similarly, Ahmed et al. \cite{ahmed2019classification} proposed three machine learning models for classifying Arabic poems into 4 types: love, Islamic, political, and social. They have used Naive Bayes, SVM, and linear support vector classifier (SVC) and had achieved an average F1-score of 0.49, 0.18, and 0.51 respectively.
Shahriar et al. \cite{shahriar2023classification} measured the performance of different deep learning models like LSTM, GRU, and CNN in the task of classifying Arabic poetry emotions. They have used 9452 poems divided into 3 classes: joy, sadness, and love. Additionally, the authors employed AraBERT model, a BERT-based model that was pretrained on Arabic text \cite{antoun2020arabert}. The fine-tuned model has achieved an F1-score of 0.77 which is significantly higher compared to the other deep learning models used in the same study that have achieved an F1-score between 0.53 and 0.62.

\subsection{Arabic Language Models}

Looking specifically at Arabic language models, multiple models have been proposed in literature since the introduction of Transformers by Vaswani et al. \cite{vaswani2017attention}. AraBERT\footnote{Available on Huggingface platform \url{https://huggingface.co/aubmindlab/bert-base-arabert}}, introduced by Antoun et al. \cite{antoun2020arabert}, is a BERT-based language model that was pretrained on a large Arabic language corpus. The 24 GB dataset is composed of Arabic news articles obtained from two publicly available corpora: 1.5 billion words Arabic corpus \cite{1.5bwords}, and OSIAN: the Open Source International Arabic News Corpus \cite{osian2019zeroual}. Additionally, the authors scraped manually more news articles from various online sources. The main reason for creating AraBERT was the need for a large language model designed specifically for the Arabic language. The authors presented two variants of AraBERT: AraBERTv0.1 and  AraBERTv1. The main difference between the two is that in AraBERTv1 the authors used \( Farasa \) \cite{abdelali2016farasa}, an Arabic text segmenter, before training the tokenizer and then tokenizing the text. Whereas in AraBERTv0.1, the Farasa segmenter was not used. When evaluating both models, AraBERTv1 outpeformed AraBERTv0.1 on six different tasks, whereas the latter achieved higher results in the remaining three tasks. Each of these variants are available in two different sizes: "base" and "large", similar to the original BERT's two sizes. AraBERT has shown impressive performance on various Arabic NLP tasks, outperforming other multilingual models that were pretrained on multiple languages including Arabic. The model has achieved state-of-the-art (SOTA) results when tested on Arabic NLP tasks such as text classification, named entity recognition, and question answering.


Similarly, Chowdhury et al. \cite{chowdhury2020qarib} proposed another BERT-based language model called QARiB\footnote{Available on Huggingface platform \url{https://huggingface.co/qarib/bert-base-qarib}}. The presented model was pretrained on text acquired from different sources including posts from Arabic news channels written in modern standard Arabic (MSA), and tweets from well known Twitter accounts that are written mostly in dialect Arabic. QARiB model has achieved higher results than AraBERT in text classification tasks on newly prepared datasets containing some text written in dialectal Arabic which shows that language models can achieve better generalization when being trained on both formal and informal text.


Abdul-Mageed et al. \cite{abdul2020arbert} introduced a new BERT-based model called ARBERT\footnote{Available on Huggingface platform \url{https://huggingface.co/UBC-NLP/ARBERT}} pretrained on 61GB of MSA text collected from Arabic Wikipedia, online free books, and publicly available corpora, mainly OSCAR \cite{oscar2019asynchronous}. The model employs a vocabulary of 100K different tokens, and was pretrained using the same configuration as the original BERT. 
The authors also introduced another model, MARBERT, that was pretrained on a different dataset composed only of tweets written in both MSA and diverse Arabic dialects. The model is designed for downstream tasks that involve dialectal Arabic. However, in this study we excluded MARBERT because most of Arabic poetry is written in classical or standard Arabic. Both models, ARBERT and MARBERT, achieved SOTA results across the majority of tasks when compared with AraBERT and other multilingual models.


Inoue et al. \cite{inoue2021camelbert} proposed a new model under the name CAMeLBERT\footnote{The model used in this study is available on Huggingface platform \url{https://huggingface.co/CAMeL-Lab/bert-base-arabic-camelbert-ca}}. In their paper they developed four different variants of CAMeLBERT: MSA, dialectal Arabic (DA), classical Arabic (CA), and mix. Each variant was pretrained on different datasets that contain a certain type of Arabic text, except for the mix variant that was pretrained on all datasets combined (167 GB).
The authors compared the new model variants with AraBERT, MARBERT, ARBERT, and other multilingual models. They showed when experimenting on tasks that involve dialectal Arabic, CAMeLBERT-DA outperformed all other models including MARBERT. Additionally, CAMeLBERT-CA outperformed all other models in the Arabic poetry task, which is the only task that is designed for evaluating language models on classical Arabic text.

In this study, we used AraBERTv1, AraBERTv0.1, ARBERT, and QARiB as comparative models to our proposed model due to their wide acceptance in literature when tackling various Arabic NLP problems \cite{alammary2022bert}. Also, we employed CAMeLBERT-CA model in this study for being mainly designed to tackle tasks that involve classical Arabic text.

\section{Transformers}\label{sec3}

\begin{figure}[h]
    \centering
    \includegraphics[width=0.5\textwidth]{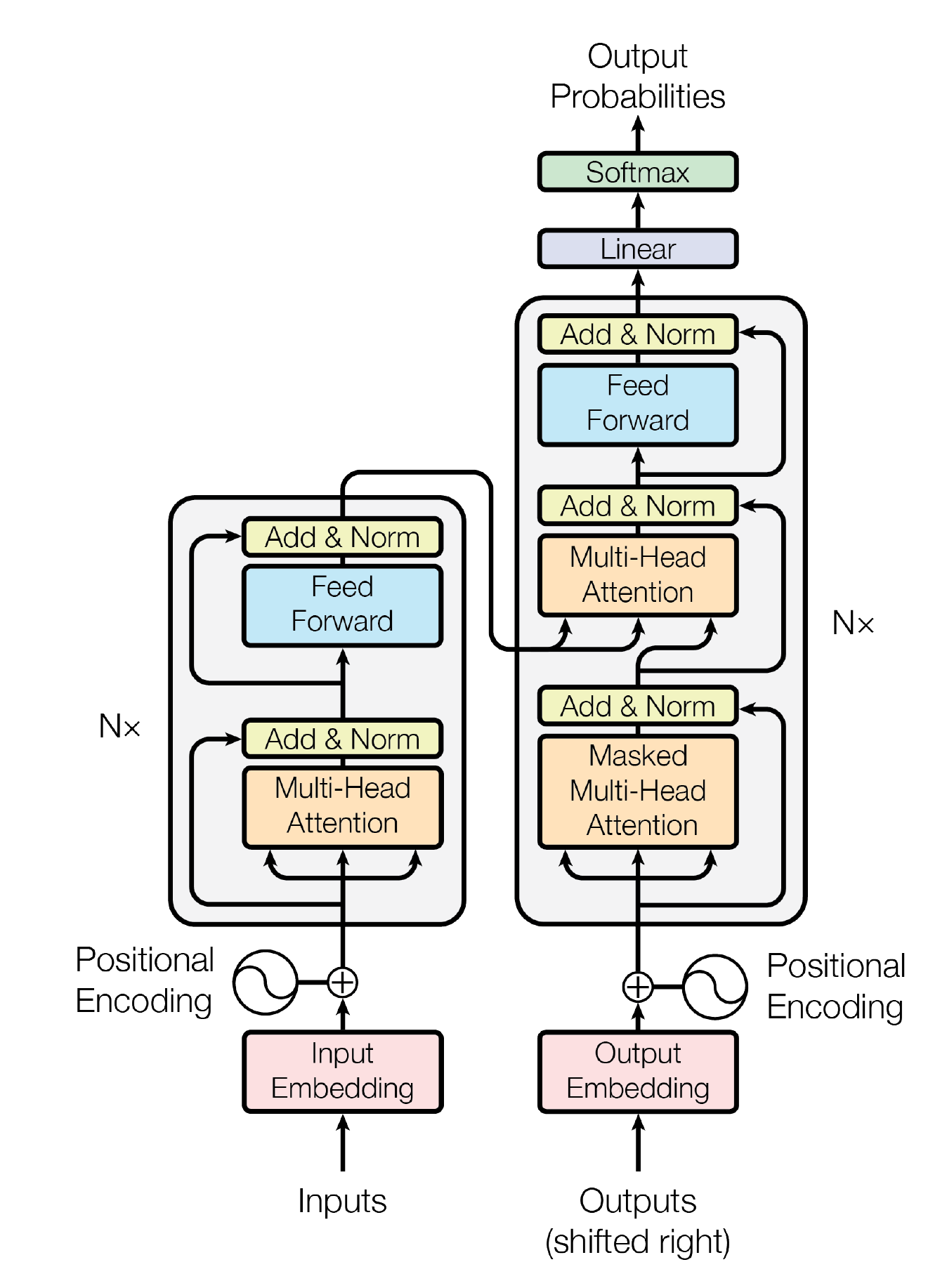}
    \caption{Overview of the transformer model architecture \cite{vaswani2017attention}}
    \label{fig1}
\end{figure}

Transformers, introduced by Vaswani et al. \cite{vaswani2017attention}, are the building block for all modern language models. They primarily utilize a mechanism called self-attention to measure the significance of different parts in the input sequence to each other. Figure \ref{fig1} shows the general architecture of Transformer model.
Given a sequence of input vectors, the self-attention mechanism computes a weighted sum of these vectors using attention scores. The core components of the self-attention mechanism are the \textit{query} (Q), \textit{key} (K), and \textit{value} (V) matrices. These are derived from the input vectors.

{}

The attention scores are computed as:
\begin{equation}
\text{Attention}(Q, K, V) = \text{softmax}\left(\frac{QK^T}{\sqrt{d_k}}\right) V
\end{equation}
where $d_k$ is the dimension of the key vectors.

While the self-attention mechanism allows the model to focus on different parts of the input, the multi-head attention mechanism allows the model to focus on different parts in different representation subspaces of Q, K, and V matrices. Essentially, it runs the self-attention mechanism multiple times in parallel, each with different learned linear projections of the original Q, K, and V.

Given $h$ different sets of Q, K, and V matrices, the multi-head attention is computed as:
\begin{equation}
\text{MultiHead}(Q, K, V) = \text{Concat}(\text{head}_1, \ldots, \text{head}_h) W_O
\end{equation}
where each head is computed as:
\begin{equation}
\text{head}_i = \text{Attention}(Q W_{Qi}, K W_{Ki}, V W_{Vi})
\end{equation}
and $W_{Qi}$, $W_{Ki}$, $W_{Vi}$, and $W_O$ are the parameter matrices.

However, since Transformers inherently lack a sense of order or position, the authors also proposed another mechanism called "positional encoding" that can give the model information about the position of words in a sequence, since all words or tokens are being processed in parallel. To address this, positional encodings are added to the embeddings at the bottoms of the encoder and decoder stacks. The positional encodings have the same dimension as the input embeddings, allowing them to be summed. The word's position \( p \) and each dimension \( i \) of the word embedding, the positional encoding is defined as:

\begin{equation}
PE_{(p, 2i)} = \sin\left(\frac{p}{{10000^{2i/d}}}\right)
\end{equation}

\begin{equation}
PE_{(p, 2i+1)} = \cos\left(\frac{p}{{10000^{2i/d}}}\right)
\end{equation}

Where \( d \) is the dimension of the embeddings. These sinusoidal functions were chosen because they can be easily learned if needed, and they allow the model to interpolate positions of words in long sequences.

The original Transformer model follows the encoder-decoder structure, where the encoder processes the input sequence and the decoder generates the output sequence. This architecture makes the original Transformer model particularly suitable for text-to-text tasks such as machine translation and paraphrase generation.

\begin{figure}[]
    \centering
    \includegraphics[width=0.5\linewidth]{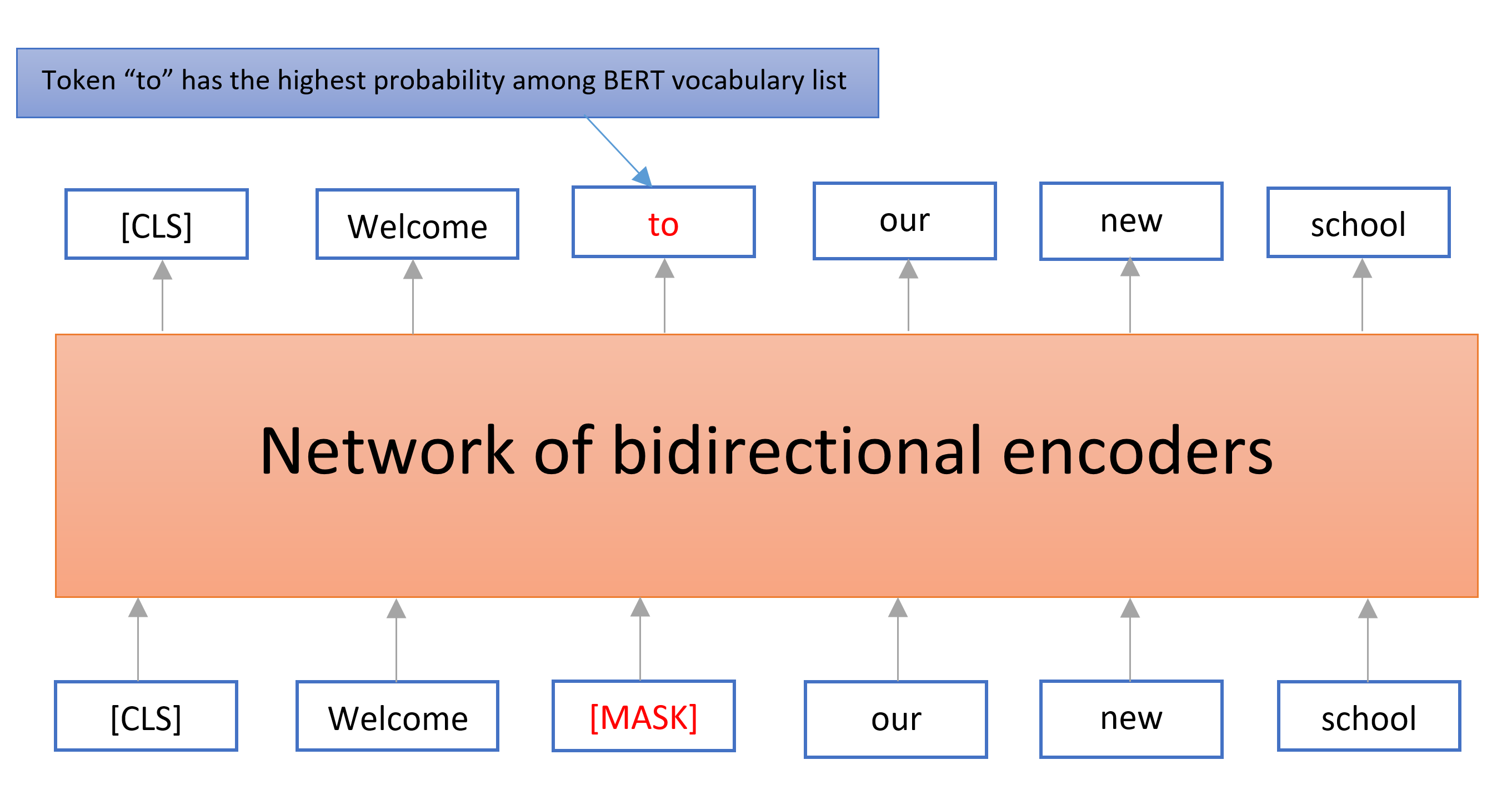}
    \caption{An example of masked language model (MLM) training objective}
    \label{fig2}
\end{figure}

Building upon the Transformer's architecture, the Bidirectional Encoder Representations from Transformers (BERT) has brought about significant advancements in natural language processing (NLP) \cite{devlin2018bert}. BERT is an encoder-only Transformer that analyzes and processes input text bidirectionally, unlike the original encoder-decoder Transformer model that reads text sequentially. One of BERT's capabilities is the ability to grasp the complete context of a word by considering its surrounding words. This is achieved through the "masked language model" (MLM) training objective, which randomly masks a percentage of input tokens and then asks the model to predict them based on the context provided by the other unmasked tokens. Figure \ref{fig2} shows an example of the MLM training objective. This bidirectional approach allows BERT to accurately comprehend the context and the meaning of each word in a sentence, especially when dealing with words that have different meanings based on their usage and surrounding words.

{}

\begin{figure}[]
    \centering
    \includegraphics[width=0.5\linewidth]{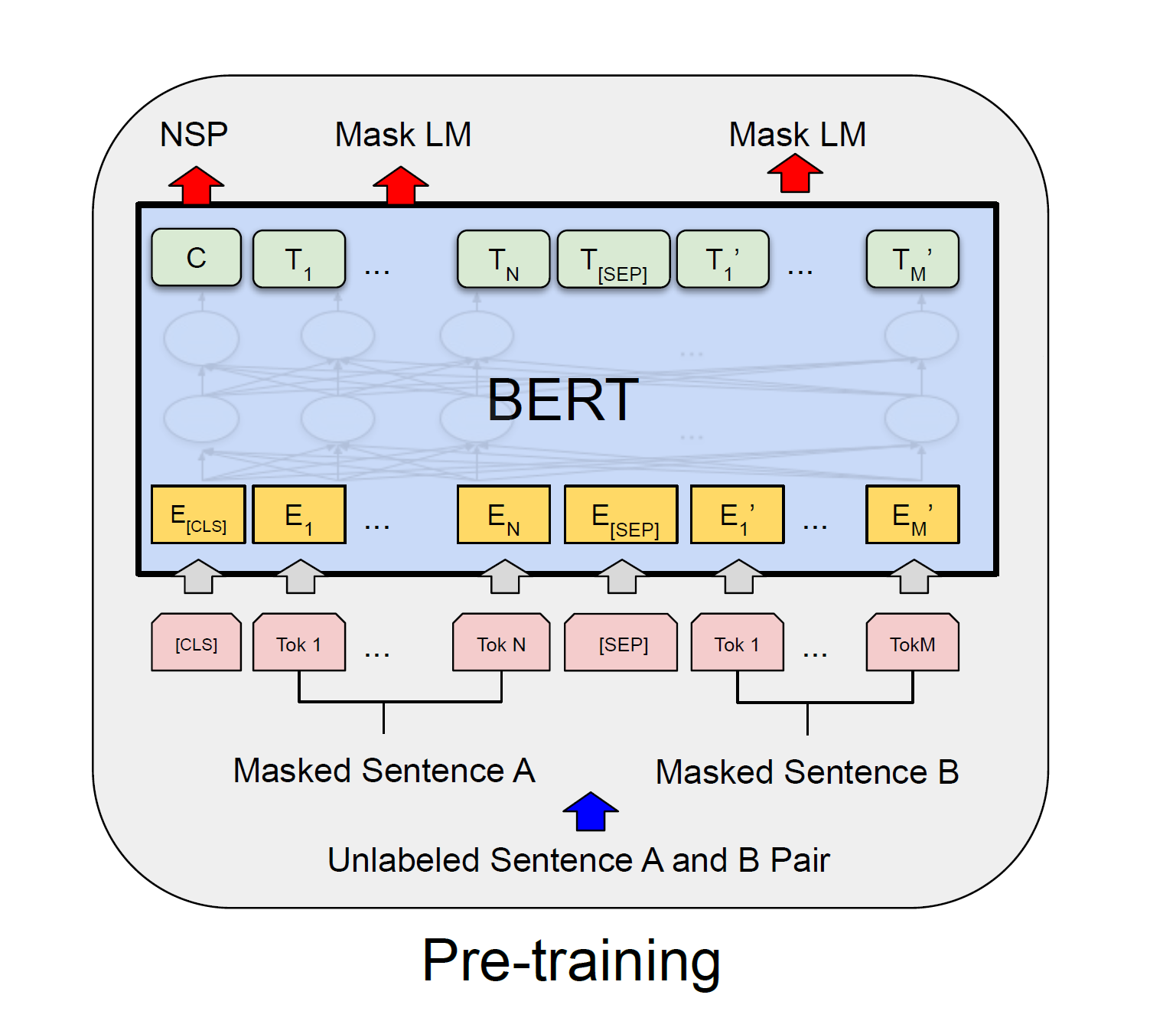}
    \caption{An illustration of BERT pretraining process \cite{devlin2018bert}.}
    \label{fig3}
\end{figure}

Additionally, BERT excels in transfer learning, which enables it to apply previously learned knowledge to different NLP tasks. Once the model has been pretrained using a large amount of text, it can be further refined by adding just one extra output layer. This allows for generating models for tasks including question answering and language inference without the need for significant modifications to the model architecture, or the need for re-traing the model for scratch. This adaptability makes BERT highly versatile and efficient, ensuring high performance across a wide range of NLP tasks. Figure \ref{fig3} shows the pretraining process of BERT language model.

The authors of BERT model have developed two different sizes of BERT: BERT-base and BERT-large. BERT-base is a model with 12 transformer blocks (layers), 768 hidden units (output vector size), and 12 attention heads for each layer, resulting in a total of 110 million parameters. BERT-large is a much larger model with 24 layers, 1024 hidden units, and 16 attention heads. It contains a total of 340 million parameters, which is 3 times larger than the base model. Both variants have been pretrained on the same dataset, but due to its larger size and complexity, BERT-large generally achieves better performance on different NLP tasks. However, it requires more computational resources and longer training time.

{}

\section{Proposed Model: AraPoemBERT}\label{sec4}

In this section we describe the proposed model architecture, and the dataset used in the study.

\subsection{Model Architecture}

\begin{figure}[]
    \centering
    \includegraphics[width=0.6\linewidth]{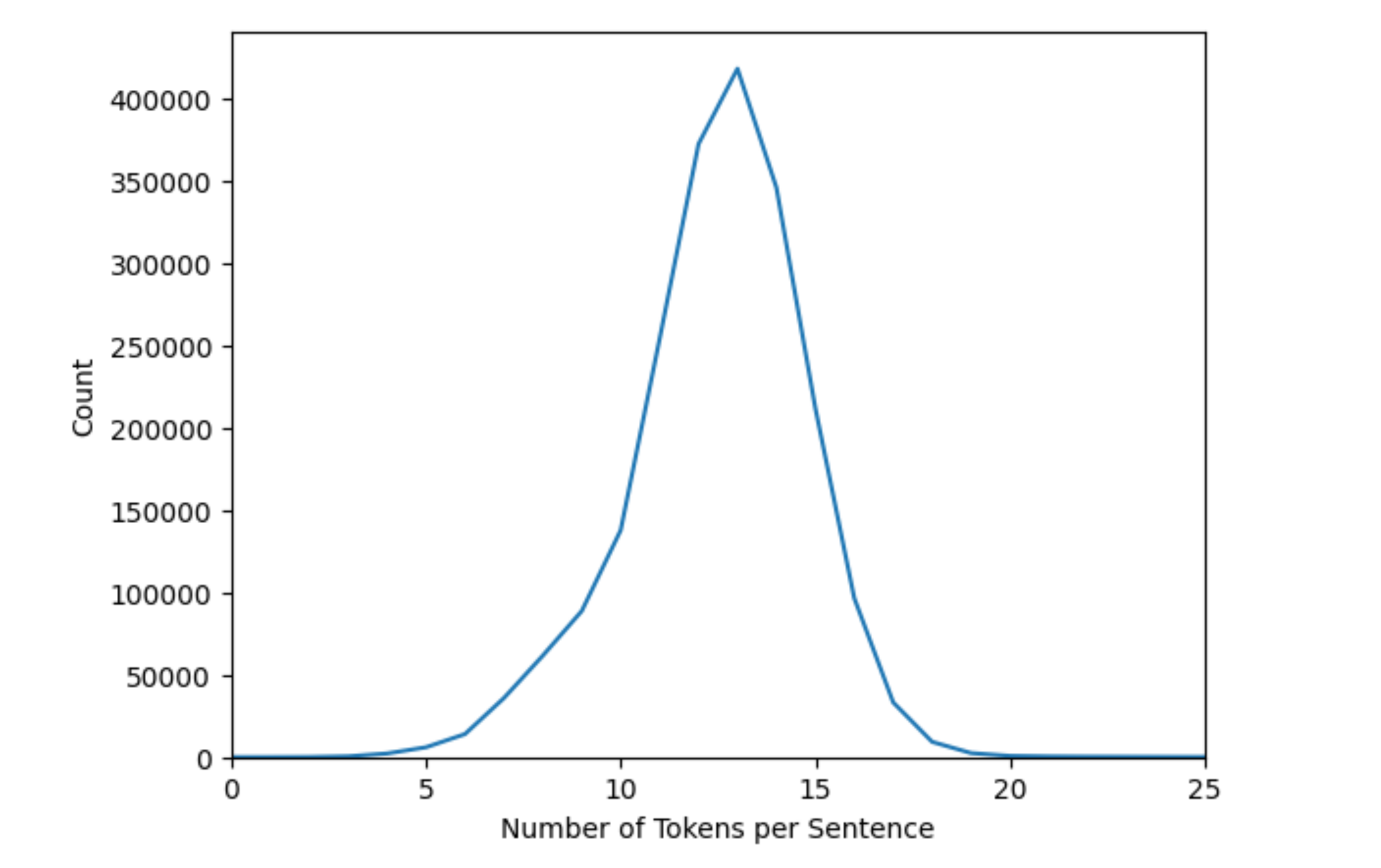}
    \caption{The distribution of number of tokens per verse in AraPoems dataset}
    \label{fig4}
\end{figure}

Building upon the success of AraBERT and many other Arabic language models, we developed AraPoemBERT, a BERT-based model that was pretrained from scratch exclusively on Arabic poetry text. The model follows the same architecture as the original BERT model in terms of the number of attention heads (12 attention heads per layer) and the size of the hidden layer (768 units). Also, we used wordpiece tokenizer \cite{wordpiece} similar to the original BERT model. However, AraPoemBERT contains 10 encoder layers, compared to 12 layers in BERT-base, and the vocabulary size of our model was set to 50,000, allowing it to capture a wide range of words and expressions found in Arabic poetry. Finally, the maximum sequence length is set to 32 tokens per sequence. The main reason for limiting the sequence length in AraPoemBERT to such a small number is due to the average length of poems' verses, where the majority of verses can be fully stored within a 32-token sequence. Figure \ref{fig4} shows that 99.3\% of sequences (a whole verse) contain between 6 and 18 tokens after tokenization. However, having a smaller sequence length does significantly reduce the model pretraining time, because it enables using larger batch sizes without causing any out-of-memory issues, even with a commodity GPU.

{}

\subsection{Dataset}
\begin{figure}[]
    \centering
    \includegraphics[width=1.2\linewidth]{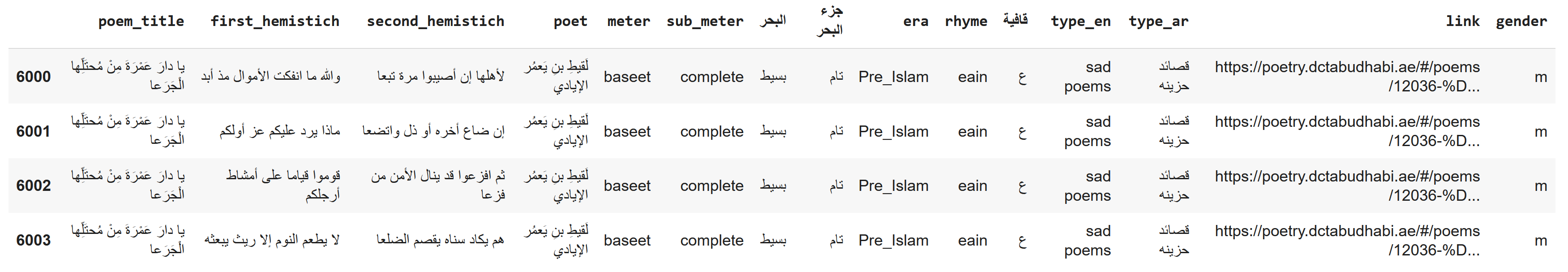}
    \caption{A sample of AraPoems dataset}
    \label{fig5}
\end{figure}

\begin{figure}[]
    \centering
    \includegraphics[width=1\linewidth]{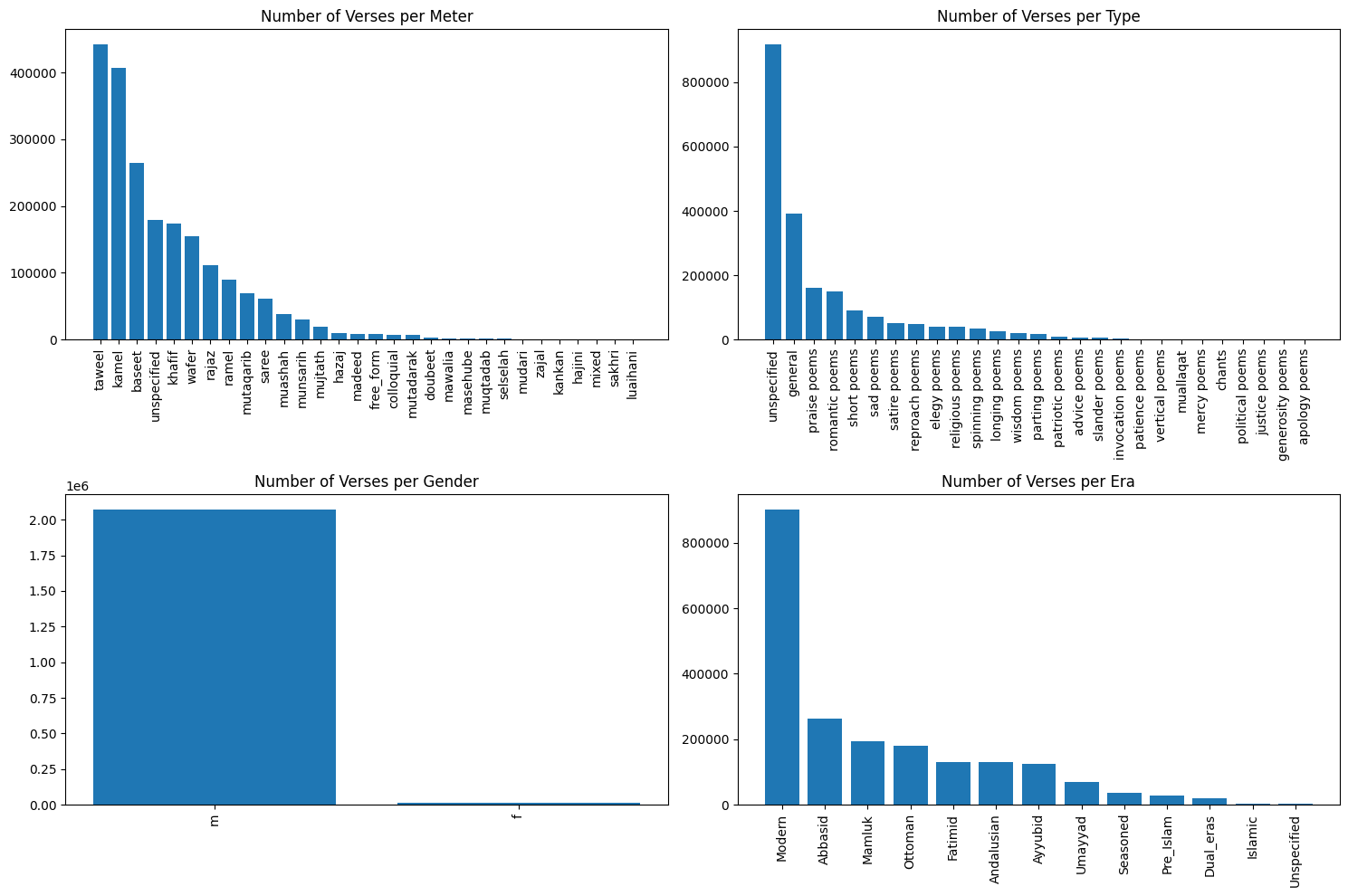}
    \caption{The distribution of verses across the categories: meter, type, gender, and era}
    \label{fig6}
\end{figure}

In this study, we used AraPoems\footnote{AraPoems dataset is available on \url{ https://doi.org/10.7910/DVN/PJPWOY}} dataset for pretraining AraPoemBERT and fine-tuning different models used in the downstream tasks. The dataset is collected from two online sources specialized in Arabic poetry, Almausua \cite{w1} and Aldiwan \cite{w2}, and it contains 2,090,907 verses associated with a variety of information such as meter, sub-meter, poet, rhyme, era, and topic. See Figure \ref{fig5} for a sample of the dataset. Figure \ref{fig6} shows the distribution of verses across different categories. Compared to the APCD dataset \cite{yousef2019learning}, the new compiled dataset contains 14\% more verses, and contains two new labels: sub-meter and the type of the poem's topic. Additionally, we translated all these information to the English language manually, and we also labeled the poets' gender based on their names. 
The dataset underwent a cleaning process which includes removing duplicate verses, and removing any irrelevant characters from the corpus such as digits, English letters, and unwanted symbols like ‘@’, ‘\#’, and ‘\$’.

{}

{}

\section{Experiments and Results}\label{sec5}

In this section, we present the text preprocessing steps, the pretraining  procedures, and the downstream tasks along with the results. All experiments were conducted on a local machine equipped with AMD Ryzen-9 7950x processor, 64GB DDR5 memory, and two GeForce RTX 4090 GPUs with 24GB memory each. The software environment was set up on Ubuntu 22.04 operating system. The Huggingface transformers library \cite{wolf2020transformers} was used in pretraining our model, in addition to downloading and fine-tuning the language models from the Huggingface hub that were used in this study. Additionally, CUDA 11.8 was used to take advantage of GPU acceleration for efficient computations.

\subsection{Text Preprocessing}

\begin{figure}[]
    \centering
    \includegraphics[width=0.6\linewidth]{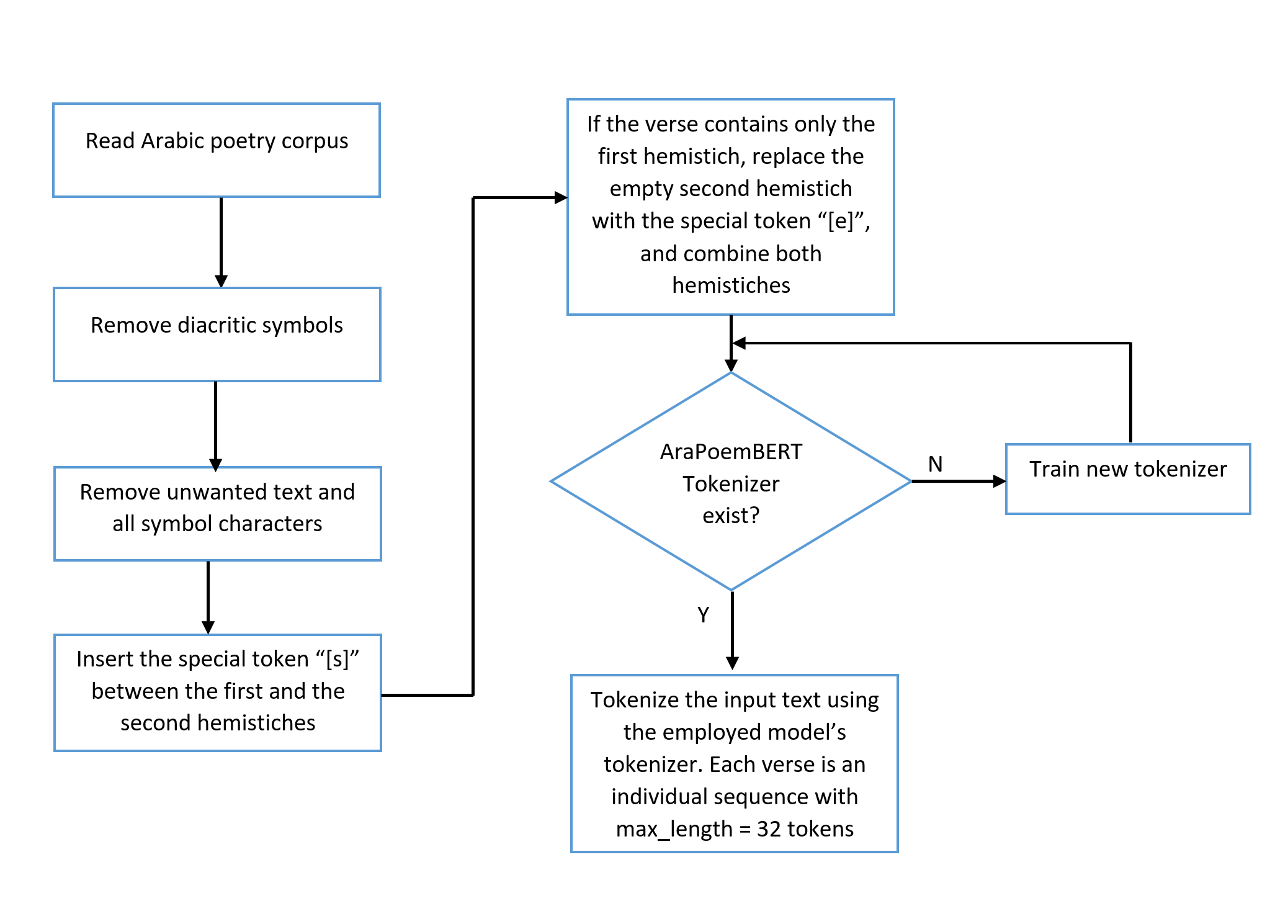}
    \caption{The algorithm structure diagram of text preprocessing}
    \label{fig7}
\end{figure}

Before using the verses' text in pretraining the model, a few preprocessing steps were required to ensure the data is in a suitable format for the model. The first step in the preprocessing phase was to remove the diacritics, which simplifies the text and reduces the pretraining time. Removing the diacritics from the text was conducted using PyArabic, a specialized Python library for manipulating and normalizing Arabic text \cite{zerrouki2010pyarabic}. The second step involves removing all symbols such as colons, brackets, and question marks from the poetry corpus. These symbols can introduce noise into the data and potentially affect the performance of the model. Moreover, due to the difference between the structure of regular text and Arabic poetry, where the verses are divided into two parts, known as hemistiches, and both hemistiches are required to form a complete sentence. And at the same time, we want to facilitate the model's understanding of the verse structure. Thus, we added two additional unique tokens: '[s]' and '[e]', where the '[s]' token is used as a separator between the first and second hemistiches in a verse, and if a verse contains only the first hemistich, the '[e]' token will be placed after the '[s]' token to represent an empty second hemistich. This approach allows the model to recognize the structure of the verses and differentiate between verses with one or two hemistiches. Figure \ref{fig7} illustrates the algorithm structure diagram of the text preprocessing stage.

{}

The final step in the preprocessing phase involves training the tokenizer and then using it whenever required to tokenize the input text. For AraPoemBERT, we employed the Huggingface implementation of the WordPiece tokenizer, which we trained on the same poetry text with a vocabulary size of 50,000 wordpieces.

\subsection{Model Pretraining}

The original BERT model was pretrained on two objective tasks: the masked language model (MLM), and next sentence prediction (NSP). In the MLM task, a certain percentage of tokens in the input sequence are masked and the model's goal is to correctly predict what these masked tokens are. The NSP task involves providing the model with two sentences and asking it to determine if they are related (from the same paragraph) or not. However, AraPoemBERT was pretrained solely on the MLM task objective which reduces the pretraining time and potentially achieves better performance in the downstream tasks, following the recommendations of the RoBERTa model's authors \cite{liu2019roberta}. Our model was pretrained by masking 15\% of the sequences' tokens, a batch size of 256, 'AdamW' \cite{loshchilov2017adamw} as the model optimizer with a learning rate of 5e-5 and weight decay equal to zero, and a dropout rate of 0.1 for all dropout layers. To reduce training time and optimize GPU memory usage, we utilized the (mixed precision) datatype "FP16" for gradient computations. With the aforementioned configurations, the model was pretrained for 800k steps (980 epochs) and it took 142 GPU hours, whereas the minimum loss reached was 2.02. In this stage, we have used all the collected poetry text which is composed of more than 2.09 million verses. The size of the text file used in pretraining the model was 182 MB and contains more than 19.22 million words or 29 million tokens after tokenization. Even though the dataset is small in size compared to other BERT-based models, Arabic poems are very distinctive and diverse, and language models in general require between 10M and 100M words to learn most of the syntactic and semantic features \cite{zhang2020you}.


\subsection{Downstream Tasks}
To demonstrate the effectiveness of AraPoemBERT model, we assessed its performance on five different downstream tasks related to Arabic poetry analysis. In this study, we used AraBERTv1, AraBERTv0.1, ARBERT, CAMeLBERT-CA, and QARiB as comparative models to our proposed model. All language models were fine-tuned using the same settings and hyperparameters values used in the pretraining process. The models were fine-tuned using 80\% of the task-related labeled data, and the remaining 20\% were reserved for validation, while ensuring that the same validation set was used when evaluating different models within the same experiment.

\subsubsection{Sentiment Analysis}

\begin{figure}[H]
\centering
\includegraphics[width=8 cm]{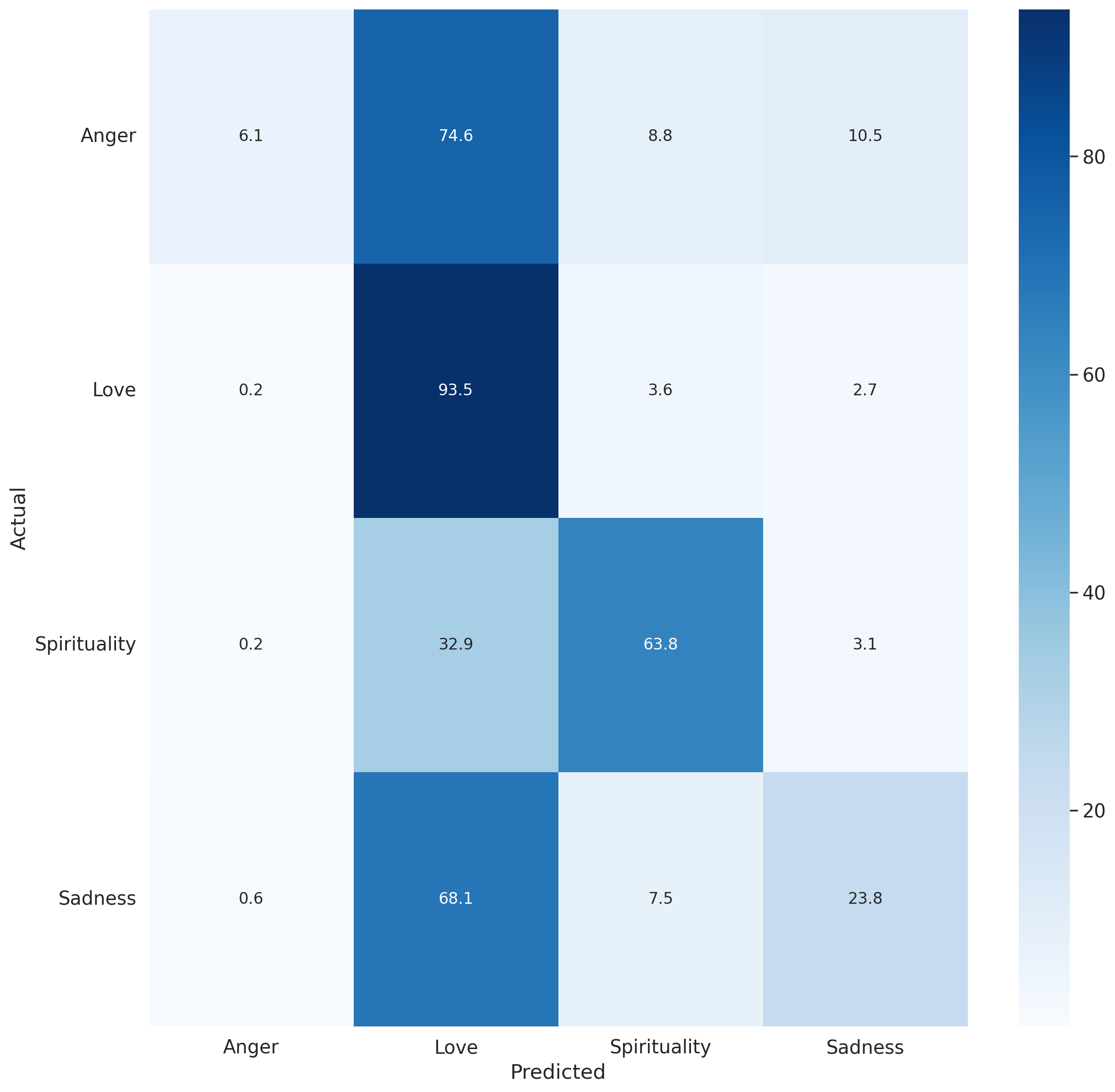}
\caption{The sentiment analysis confusion Matrix for the test set using AraPoemBERT.\label{fig8}}
\end{figure}

Many poems in the dataset are labeled with different types of topics such as: Romantic poems, Elegy poems, Invocation poems, etc. However, we grouped relevant types of poems into a single emotion, resulting in four different classes: Love, Sadness, Anger, and Spirituality. The resulting dataset labeled with these four emotions contains 21,230 poems composed of 315,877 verses.
Table \ref{tab4} shows the list of poems' types and their relevant emotions. As shown in Table \ref{tab5}, our model has outperformed all other language models with an accuracy score of 78.95\% which is significantly higher than the second score achieved by CAMeLBERT-CA model, while the lowest accuracy results are achieved by both AraBERT models (75.35\% and 75.79\%).
Table \ref{tab6} presents the classification scores achieved by AraPoemBERT for each sentiment class in the validation set. The model attained high scores in the classes "Love" and "Spirituality", while it wrongly classifies "Sadness" and "Anger" with the class "Love" as shown in the confusion matrix in Figure \ref{fig8}, which explains the low F1-score for these two classes. Overall, AraPoemBERT has achieved a significantly higher results compared to previous work by Shahriar et al. \cite{shahriar2023classification}. In their work, they targeted three emotion classes only, and achieved an accuracy of 76.5\% with a dataset composed of 9452 poems, which is less than half the number of poems used in this study. Additionally, in their work, the proposed models require processing whole poems to accurately detect the main sentiment. However, in this study, we achieved higher accuracy by merely evaluating individual verses to predict the related poem's main sentiment, which shows the significance of language models pretrained on a domain-specific text such as AraPoemBERT, or CAMeLBERT-CA which was pretrained solely on classical Arabic text.

\begin{table}[ht]
\centering
\caption{Poems types and emotions.~\label{tab4}}
\begin{tabular}{lccc}
\hline
\textbf{Poem Types} & \textbf{Sentiment} & \textbf{\#Verses} & \textbf{\#Poems} \\
\hline
Slander Poems & Anger & 5079 & 631 \\
Romantic, Parting, Longing, and Spinning Poems & Love & 226368 & 15228 \\
Religious, Invocation, and Mercy Poems & Spirituality & 43429 & 2275 \\
Elegy Poems & Sadness & 41001 & 3172 \\
Total &  & 315877 & 21230 \\
\hline
\end{tabular}
\end{table}

\begin{table}[ht]
\centering
\caption{The accuracy results achieved by various language models in the sentiment analysis task.~\label{tab5}}
\begin{tabular}{llll}
\toprule
\textbf{Name} & \textbf{Accuracy} & \textbf{Dataset Size} & \textbf{Classes} \\
\midrule
Shahriar et al. (AraBERTv0.2-base) \cite{shahriar2023classification} & 76.50\% & 9452 poems & Sad, Love, Joy \\
AraBERTv0.1-base & 75.79\% & \multirow{6}{*}{21,230 poems (315,877 verses)} & \multirow{6}{*}{Love, Sadness, Anger, Spirituality} \\
AraBERTv1-base & 75.35\% & & \\
QARiB & 76.55\% & & \\
ARBERT & 76.23\% & & \\
CAMeLBERT-CA & 77.77\% & & \\
AraPoemBERT-base (ours) & \textbf{78.95\%} & & \\
\bottomrule
\end{tabular}

\end{table}


\begin{table}[ht]
\centering
\captionsetup{width=1\textwidth}
\caption{The classification report of AraPoemBERT in the sentiment analysis task.~\label{tab6}}
\begin{tabular}{ccccc}
\hline
\textbf{Class} & \textbf{Precision} & \textbf{Recall} & \textbf{F1-Score} & \textbf{Number of Samples} \\
\hline
Anger & 0.2952 & 0.0610 & 0.1011 & 1016 \\
Love & 0.8215 & 0.9348 & 0.8745 & 45274 \\
Spirituality & 0.7018 & 0.6380 & 0.6684 & 8686 \\
Sadness & 0.5512 & 0.2384 & 0.3329 & 8200 \\
\hline
 \multicolumn{5}{c}{}  \\
Accuracy &  &  & 0.7896 & 63176 \\
Macro Avg & 0.5924 & 0.4681 & 0.4942 & 63176 \\
Weighted Avg & 0.7615 & 0.7896 & 0.7634 & 63176 \\
\hline
\end{tabular}
\end{table}

\subsubsection{Poetry Meters}

The task of classifying Arabic poetry meters has proven to be quite challenging in literature, especially when tackling a dataset composed of a large number of verses, and at the same time, multiple meters are underrepresented in the dataset. We have conducted two classification tasks; one focused on classical meters only (16 meters) to compare our model accuracy with existing work, and the second classification task includes both classical and non-classical meters (28 meters). Table \ref{tab7} shows a list of classical and non-classical meters found in the AraPoems dataset.

\begin{table}
\captionsetup{width=0.8\textwidth}
\caption{\textbf{List of Classical and Non-Classical meters}}
\label{tab7}
\setlength{\tabcolsep}{3pt}
\centering
\begin{tabular}{|c|c|c|c|c|c|}
\hline
\multicolumn{3}{|c|}{Classical Meters} & \multicolumn{3}{|c|}{Non-Classical Meters}\\
\hline
No.&
Meter&
No, of Verses&
No.&
Meter&
No, of Verses \\
\hline
1   &  Taweel   & 442763 & 17  &   Muashah& 38409 \\
2   &  Kamel   &  406908 & 18  &   Free form& 8431 \\
3   &  Baseet   &  265045 & 19  &   Colloquial& 7399 \\
4   &  Khafif   &  174105 & 20  &   Doubeet& 2937 \\
5   &  Wafer   &  154455 & 21  &   Mawalia& 1887 \\
6   &  Rajaz   &  111333 &  22  &   Masehube& 1283 \\
7   &  Ramel   & 89857 &  23  &   Selselah&  945 \\
8   &  Mutaqarib   & 69523 &  24  &   Zajal& 81 \\
9   &  Saree   &  61261 & 25  &   Kankan& 59 \\
10   &  Munsarih   &  29863 & 26  &   Hajini& 48 \\
11   &  Mujtath   &  19357 & 27  &   Sakhri& 16 \\
12   &  Hazaj   &  9166 & 28  &   Luaihani& 7 \\
13   &  Madeed   &  8775 &  &  & \\
14   &  Mutadarak   & 6631 &    &   &  \\
15   &  Muqtadab   &  949 &  &    & \\
16   &  Mudari   &   360 &  &    & \\
\hline
\multicolumn{2}{|c|}{Total} & 1850351  & \multicolumn{2}{|c|}{Total} & 61502\\
\hline
\end{tabular}
\end{table}

\begin{table}
\caption{\textbf{Poetry meters classification results achieved by various language models}}
\label{tab8}

\begin{tabular}{lllp{1.7cm}p{2.8cm}}

\toprule
Name &   Architecture & Accuracy (classical meters) & Dataset Size (No. Verses) & Accuracy (all meters) \\
\midrule
 Al-Shaibani et al. \cite{alShaibani2020meter} & 5-BiGRU layers & (14 meters only) 94.32\% & 55K & NA  \\
Yousuf et al. \cite{yousef2019learning} & 7-BiLSTM & 94.11\% & 1,72M & NA \\
Abandah et al. \cite{abandah2022classifying}& 4-BiLSTM & 97.27\% & 1.62M & NA \\
AraBERTv01-base & 12L12H & 98.31\% & 1.85M & 96.92\% \\
 AraBERTv1-base & 12L12H & 98.25\% & 1.85M & 96.87\% \\
 QARiB & 12L12H & 98.26\% & 1.85M & 96.88\% \\
 ARBERT & 12L12H & 98.03\% & 1.85M & 96.59\% \\
CAMeLBERT-CA & 12L12H & 98.50\% & 1.85M & 97.23\% \\
AraPoemBERT-base (ours) & 10L12H & \textbf{99.03\%} & 1.85M & \textbf{97.82\%} \\
\bottomrule
\end{tabular}
\end{table}


In Table \ref{tab8}, we compare our model with other language model and machine learning models presented in literature. AraPoemBERT has achieved the highest accuracy score in both tasks outperforming other models including proposed approaches from previous work. Regarding the classification task that targets all poetry meters, which includes an additional 12 non-classical meters, we compared our model with other language models only, because to the best of our knowledge there were no published results in literature that cover this area.

\begin{figure}[]
    \centering
    \includegraphics[width=1\linewidth]{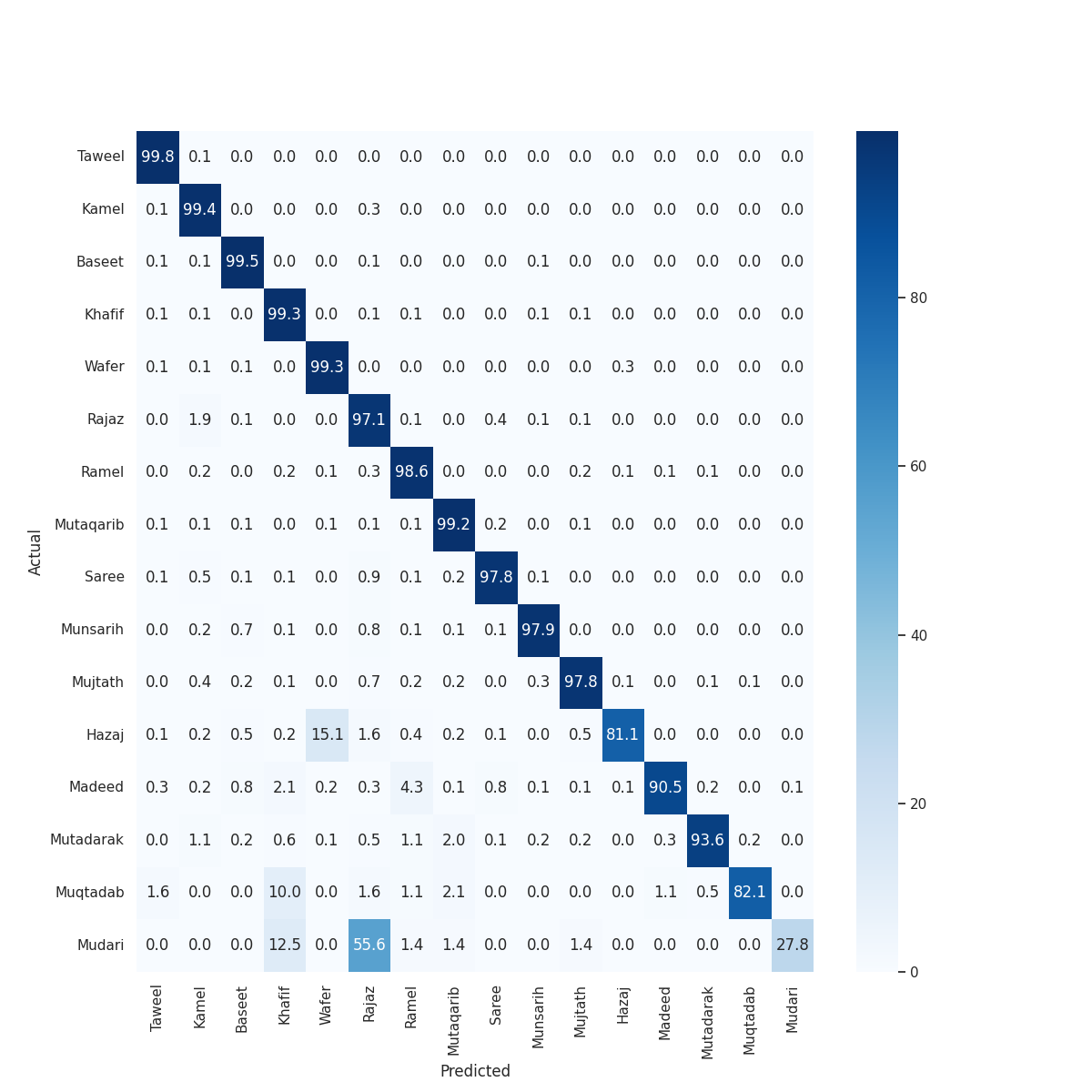}
    \caption{
    The confusion matrix for the test set in the task of classifying classical meters using AraPoemBERT.
    }
    \label{fig9}
\end{figure}

Table \ref{tab9} shows the prediction results of our model in classifying the 16 classical meters. These results include precision, recall and F1 scores for each class, in addition to the number of verses for each meter in the validation set. The model has successfully achieved an accuracy of 99\% in six different meters (1.Taweel, 2.Kamel, 3.Baseet, 4.Khafif, 5.Wafer, and 8.Mutaqarib), setting a new state-of-the-art result that has not been reported in previous work except for 'Taweel' meter only. The model also achieved an accuracy score of 97\% and 98\% in another five meters (6.Rajaz, 7.Ramel, 9.Saree, 10.Munsarih, and 11.Mujtath) despite the fact that the total number of verses related to 'Munsarih' and 'Mujtath' meters are relatively small compared to other meters. Similarly, the number of samples for '13.Madeed' and '14.Mutadarak' meters are also small but the model has achieved an accuracy score of 93.83\% and 94.41\% respectively. However, the accuracy scores of '12.Hazaj' and '15.Muqtadab' are a little lower, because the model wrongly classifies 15\% of 'Hazaj' verses as 'Rajaz', and 10\% 'Muqtadab' verses as 'Khafif' as shown in the confusion matrix in Figure \ref{fig9}. Lastly, the accuracy score of '16.Mudari' meter is the lowest, this is due to the small number of verses for this meter, and the model wrongly classified 55.6\% of this meter's samples as 'Rajaz' as shown in the confusion matrix.

\begin{table}
\captionsetup{width=1\textwidth}
\caption{\textbf{Prediction results of AraPoemBERT for classifying classical meters on an independent validation set}}
\label{tab9}
\setlength{\tabcolsep}{3pt}
\centering
\begin{tabular}{|c|c|c|c|c|c|}
\hline

Label & Meter & Precision & Recall & F1-score & No. of Verses \\
\hline

1 & Taweel & 0.9972 & 0.9977 & 0.9975 &  88553\\
2 & Kamel & 0.9906 & 0.9937 & 0.9921 & 81382 \\
3& Baseet & 0.9961 & 0.995 & 0.9955 & 53009 \\
4& Khafif & 0.9942 & 0.9935 & 0.9938 & 34821 \\
5& Wafer & 0.9876 & 0.9935 & 0.9905 & 30891 \\
6& Rajaz & 0.9709 & 0.9709 & 0.9709 & 22267 \\
7& Ramel & 0.9857 & 0.9857 & 0.9857 &  17971\\
8& Mutaqarib & 0.9916 & 0.9919 & 0.9918 & 13905 \\
9& Saree & 0.9839 & 0.9782 & 0.9811 & 12252 \\
10& Munsarih & 0.9737 & 0.9787 & 0.9762 & 5973 \\
11& Mujtath & 0.9653 & 0.9775 & 0.9714 &  3871\\
12& Hazaj & 0.9112 & 0.8112 & 0.8583 & 1833 \\
13& Madeed & 0.9742 & 0.9048 & 0.9383 &  1755\\
14& Mutadarak & 0.9524 & 0.9359 & 0.9441 & 1326 \\
15& Muqtadab & 0.8914 & 0.8211 & 0.8548 &  190\\
16& Mudari & 0.7143 & 0.2778 & 0.4 & 72 \\

\hline
\end{tabular}
\end{table}

{}

\begin{figure}[]
    \centering
    \includegraphics[width=1\linewidth]{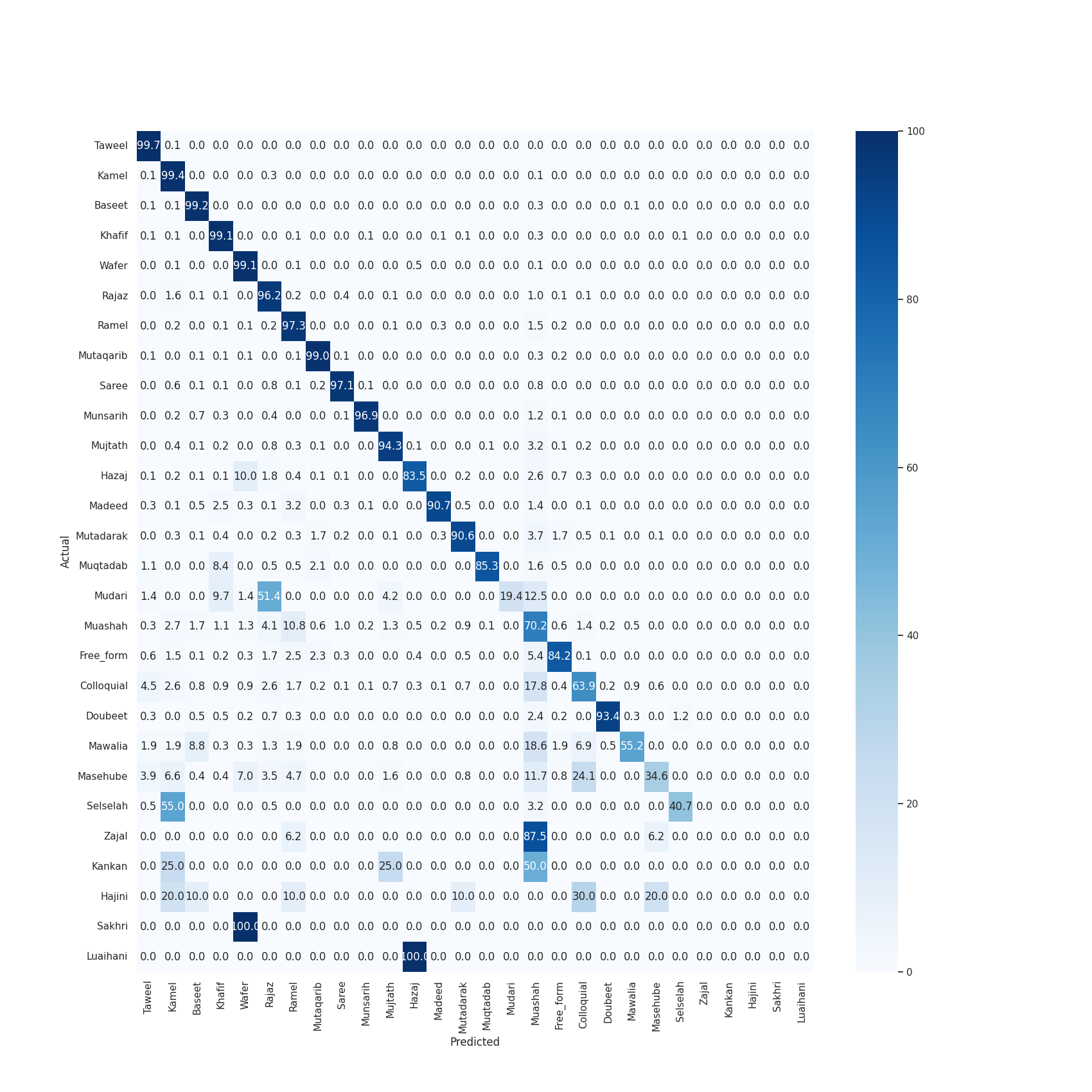}
    \caption{
    The confusion matrix for the test set in task of classifying all meters (both classical and non-classical meters) using AraPoemBERT.
    }
    \label{fig10}
\end{figure}

The prediction results of AraPoemBERT in classifying all meters (16 classical and 12 non-classical) are shown in Table \ref{tab10}. The model has achieved similar accuracy scores for the classical meters even when including the other meters in the classification task. Notably, the model achieved a score of 93.92\% with (20.Doubeet) meter, and an accuracy score between 70\% and 85\% in '17.Muashah', '18.Free form', and '19.Colloquial' meters. However, the model achieved an accuracy score between 49\% and 61\% in the meters '21.Mawalia', '22.Masehube', and '23.Selselah' because of the small number of samples for these meters, and the model wrongly classifies them as different meters such as 'Muashah', 'Colloquial', and 'Kamel' as shown in the confusion matrix in Figure \ref{fig10}. Finally, the model couldn't detect and correctly classify any of the last five meters (24.Zajal, 25.Kankan, 26.Hajini, 27.Sakhri, and 28.Luaihani) because the number of samples for each of these meters is less than 100 verses and detecting them by the model currently seems unfeasible.

\begin{table}
\captionsetup{width=1\textwidth}
\caption{\textbf{Prediction results of AraPoemBERT for classifying all meters on an independent validation set}}
\label{tab10}
\setlength{\tabcolsep}{3pt}
\centering
\begin{tabular}{|c|c|c|c|c|c|}
\hline

No. & Meter & Precision & Recall & F1-score & Num. of Verses \\
\hline

1 & Taweel & 0.9963 & 0.9971 & 0.9967 &  88553\\
2 & Kamel & 0.9863 & 0.9937 & 0.99 & 81382 \\
3& Baseet & 0.9933	 & 0.9916 & 0.9925 & 53009 \\
4& Khafif & 0.9908 & 0.9909 & 0.9908 & 34821 \\
5& Wafer & 0.986 & 0.9913 & 0.9886 & 30891 \\
6& Rajaz & 0.958 & 0.962 & 0.96 & 22267 \\
7& Ramel & 0.9384 & 0.9725 & 0.9552 &  17971\\
8& Mutaqarib & 0.9863 & 0.99 & 0.9882 & 13905 \\
9& Saree & 0.9793 & 0.971 & 0.9752 & 12252 \\
10& Munsarih & 0.982 & 0.969 & 0.9755 & 5973 \\
11& Mujtath & 0.9513 & 0.9432 & 0.9472 &  3871\\
12& Hazaj & 0.8813 & 0.8347 & 0.8574 & 1833 \\
13& Madeed & 0.9398 & 0.9066 & 0.9229 &  1755\\
14& Mutadarak & 0.8916 & 0.9057 & 0.8986 & 1326 \\
15& Muqtadab & 0.9 & 0.8526 & 0.8757 &  190\\
16& Mudari & 0.875 & 0.1944 & 0.3182 & 72 \\
 17& Muashah & 0.7505 & 0.7024 & 0.7257 & 7682 \\
 18& Free form & 0.857 & 0.8422 & 0.8495 &  1686\\
 19& Colloquial & 0.7773 & 0.6392 & 0.7015 & 1480 \\
 20& Doubeet & 0.9448 & 0.9336 & 0.9392 & 587 \\
 21& Mawalia & 0.698 & 0.5517 & 0.6163 &  377 \\
 22& Masehube & 0.8558 & 0.3463 & 0.4931 & 257 \\
 23& Selselah & 0.6875 & 0.4074 & 0.5116 & 189 \\
 24& Zajal & 0 & 0 & 0 & 16 \\
 25& Kankan & 0 & 0 & 0 & 12 \\
 26& Hajini & 0 & 0 & 0 & 10 \\
 27& Sakhri & 0 & 0 & 0 & 3 \\
 28& Luaihani & 0 & 0 & 0 & 1 \\

\hline
\end{tabular}
\end{table}

{}


\subsubsection{Poetry Sub-Meters}

To further expand the problem of classifying classical meters, we include what is called meter's variants or sub-meters to the poetry meters classification task. The majority of verses in the newly compiled dataset are labeled with a specific meter and a meter's variant. However, in order to reduce the classification problem complexity, we combined both meters and their variants to form a new set of labels. For example, "Khafif" meter comes in two variations: "Complete" and "Majzuu", thus, their combination will result in two different classes: "Khafif Complete" and "Khafif Majzuu". In this study, we will refer to the combination of meters and their variants as 'sub-meters'. After combining the meters and their variants into combined classes, we ended up with a total of 33 different sub-meters. However, we excluded sub-meters classes from the classification task if they contain about 100 verses or less, which resulted in the removal of seven sub-meters from the experiment. See Table \ref{tab11} for the list of removed sub-meters. The remaining 25 sub-meters, which account for approximately 88.48\% of all verses in the dataset, will be the focus of the classification task. Table \ref{tab12} shows the list of target sub-meters and the number of verses and poems for each sub-meter. Table \ref{tab13} shows the accuracy results of all six models used in the study, with AraPoemBERT achieving the highest accuracy score (97.79\%). 

The classification report of the validation set for AraPoemBERT in Table \ref{tab14} shows that the model has achieved an F1-score over 0.98 in all classes with the 'Complete' variant, except for 'Mutadarak Complete' and 'Rajaz Complete' due to the low number of samples for these two sub-meters. Also, the F1-score was below 0.4 for the sub-meters that contain a low number of verses that are less than 1200 in the validation set. Figure 2 shows the confusion matrix of the validation set using AraPoemBERT in the sub-meters classification task. To the best of our knowledge, this is the first study in literature that directly focuses on the problem of classifying meters variants. However, compared to the results of classifying classical meters in previous work, the accuracy of classifying sub-meters using AraPoemBERT is better than the results reported by Abandah et al. \cite{abandah2022classifying}(97.79\% VS 97.27\%) even after increasing the complexity of the problem, but significantly lower when compared with the best accuracy score reported in the previous task when plainly classifying classical meters without any consideration to their variants (97.79\% VS 99.03\%).

\begin{table}[ht]
\centering
\captionsetup{width=0.6\textwidth}
\caption{List of excluded sub-meters.~\label{tab11}}
\begin{tabular}{|c|c|}
\hline
\textbf{Sub-Meter} & \textbf{No. of Verses} \\
\hline
Kamel Maktuu & 5 \\
\hline
Mutadarak Manhuk & 7 \\
\hline
Baseet Mashture & 12 \\
\hline
Munsarih Manhuk & 37 \\
\hline
Mutaqarib Majzuu & 40 \\
\hline
Mutadarak Majzuu & 45 \\
\hline
Rajaz Manhuk & 77 \\
\hline
Baseet Majzuu & 101 \\
\hline
\end{tabular}
\end{table}

\begin{table}[ht]
\centering
\caption{List of target sub-meters included in the study.~\label{tab12}}
\begin{tabular}{|c|c|c|c|}
\hline
\textbf{No.} & \textbf{Sub-Meter} & \textbf{No. of Verses} & \textbf{No. of Poems} \\
\hline
1 & Baseet Complete & 257226 & 22442 \\
2 & Baseet Mukhala & 1986 & 300 \\
3 & Hazaj Majzuu & 9027 & 837 \\
4 & Kamel Ahuth & 1634 & 227 \\
5 & Kamel Complete & 390976 & 28945 \\
6 & Kamel Majzuu & 5592 & 672 \\
7 & Khafif Complete & 169246 & 13129 \\
8 & Khafif Majzuu & 1135 & 130 \\
9 & Madeed Majzuu & 8459 & 780 \\
10 & Mudari Majzuu & 358 & 31 \\
11 & Mujtath Majzuu & 18820 & 2463 \\
12 & Munsarih Complete & 29054 & 3382 \\
13 & Muqtadab Majzuu & 945 & 65 \\
14 & Mutadarak Complete & 6205 & 275 \\
15 & Mutadarak Mashture & 168 & 7 \\
16 & Mutaqarib Complete & 68139 & 6724 \\
17 & Rajaz Complete & 7586 & 384 \\
18 & Rajaz Majzuu & 100358 & 6871 \\
19 & Rajaz Mashture & 1351 & 183 \\
20 & Ramel Complete & 84372 & 6217 \\
21 & Ramel Majzuu & 3651 & 393 \\
22 & Saree Complete & 60051 & 8357 \\
23 & Taweel Complete & 433142 & 36970 \\
24 & Wafer Complete & 150115 & 14658 \\
25 & Wafer Majzuu & 745 & 89 \\
\hline
\end{tabular}
\end{table}

\begin{table}[ht]
\centering
\captionsetup{width=1\textwidth}
\caption{The accuracy results in three different classification tasks: sub-meter, poet's gender, and poetry rhymes.~\label{tab13}}

\begin{tabular}{lcccc}
\toprule
\textbf{Name} & \textbf{Sub-Meter} & \textbf{Gender} & \textbf{Rhyme} \\
\midrule
AraBERTv0.1-base & 97.16\% & 99.28\% & 97.57\% \\
AraBERTv1-base & 97.15\% & 99.29\% & 97.33\% \\
QARiB & 97.01\% & 99.32\% & 97.58\% \\
ARBERT & 96.86\% & 99.28\% & 97.40\% \\
CAMeLBERT-CA & 97.46\% & 99.28\% & \textbf{97.76\%} \\
AraPoemBERT-base (ours) & \textbf{97.79\%} & \textbf{99.34\%} & 97.73\% \\
\bottomrule
\end{tabular}
\end{table}


\begin{table}[ht]
\centering
\caption{Sub-meters classification report using AraPoemBERT model.~\label{tab14}}
\begin{tabular}{|c|c|c|c|c|c|}
\hline
\textbf{Label} & \textbf{Class} & \textbf{Precision} & \textbf{Recall} & \textbf{F1-score} & \textbf{Number of Samples} \\
\hline
0 & Baseet complete & 0.9884 & 0.9959 & 0.9921 & 52585 \\
1 & Baseet mukhala & 0.5484 & 0.0423 & 0.0785 & 402 \\
2 & Hazaj majzuu & 0.8850 & 0.8603 & 0.8725 & 1833 \\
3 & Kamel ahuth & 0.5000 & 0.1206 & 0.1943 & 340 \\
4 & Kamel complete & 0.9769 & 0.9914 & 0.9841 & 79902 \\
5 & Kamel majzuu & 0.5465 & 0.1238 & 0.2019 & 1139 \\
6 & Khafif complete & 0.9874 & 0.9945 & 0.9909 & 34589 \\
7 & Khafif majzuu & 0.3485 & 0.0991 & 0.1544 & 232 \\
8 & Madeed majzuu & 0.9580 & 0.9225 & 0.9399 & 1755 \\
9 & Mudari majzuu & 0.8462 & 0.3056 & 0.4490 & 72 \\
10 & Mujtath majzuu & 0.9783 & 0.9773 & 0.9778 & 3871 \\
11 & Munsarih complete & 0.9831 & 0.9780 & 0.9806 & 5965 \\
12 & Muqtadab majzuu & 0.9458 & 0.8263 & 0.8820 & 190 \\
13 & Mutadarak complete & 0.9501 & 0.9353 & 0.9426 & 1282 \\
14 & Mutadarak mashture & 0.5238 & 0.3333 & 0.4074 & 33 \\
15 & Mutaqarib complete & 0.9910 & 0.9921 & 0.9915 & 13897 \\
16 & Rajaz complete & 0.4725 & 0.5063 & 0.4888 & 1580 \\
17 & Rajaz majzuu & 0.9236 & 0.9266 & 0.9251 & 20396 \\
18 & Rajaz mashture & 0.4351 & 0.2065 & 0.2801 & 276 \\
19 & Ramel complete & 0.9511 & 0.9846 & 0.9675 & 17226 \\
20 & Ramel majzuu & 0.5743 & 0.1557 & 0.2450 & 745 \\
21 & Saree complete & 0.9814 & 0.9830 & 0.9822 & 12252 \\
22 & Taweel complete & 0.9977 & 0.9975 & 0.9976 & 88553 \\
23 & Wafer complete & 0.9872 & 0.9909 & 0.9891 & 30741 \\
24 & Wafer majzuu & 0.4400 & 0.0733 & 0.1257 & 150 \\
\hline
 \multicolumn{3}{|c}{Accuracy}              &           &0.9780 &   370006 \\
\hline
\end{tabular}
\end{table}


\subsubsection{Poet's Gender}

The dataset originally did not contain any information regarding poets' gender. Thus, we manually annotated the poets' gender based on their names. The dataset contains a total of 5,383 poets, of which 5,023 are males and 360 poets are females. Almost all verses in the dataset, specifically 2,087,557 verses, are associated with known poets. Table \ref{tab13} shows the overall accuracy results for all the models. Table \ref{tab15} presents the classification report for AraPoemBERT, which shows that the model has achieved an F1-score of 0.2246 for the Female class even though it is extremely underrepresented compared to the Male class, and 99.12\% weighted average accuracy for both classes.

\begin{table}[ht]
\centering
\captionsetup{width=1\textwidth}
\caption{Poet's gender classification report using AraPoemBERT model. ~\label{tab15}}
\begin{tabular}{ccccc}
\hline
\textbf{Class} & \textbf{Precision} & \textbf{Recall} & \textbf{F1-Score} & \textbf{Number of Samples} \\
\hline
Female & 0.7582 & 0.1318 & 0.2246 & 2996 \\
Male & 0.9938 & 0.9997 & 0.9967 & 414516 \\
\hline
 &  &  &  & \\
Accuracy &  &  & 0.9935 & 417512 \\
Macro Avg & 0.8760 & 0.5658 & 0.6107 & 417512 \\
Weighted Avg & 0.9921 & 0.9935 & 0.9912 & 417512 \\
\hline
\end{tabular}
\end{table}


\subsubsection{Poem's Rhyme}

In the task of rhyme classification, the verses in the dataset are labeled with 31 different rhymes. These rhymes include all 28 Arabic letters, in addition to the rhymes: Laa, Taa Marbutah, and Waw Hamza which are variants of the letters Lam, Taa, and Alif, respectively, but they are written differently and have slightly different sounds. This classification task aims to accurately identify the rhyme of each verse, providing further insight into the structure and style of the poem.

\begin{figure}[]
\includegraphics[scale=0.25]{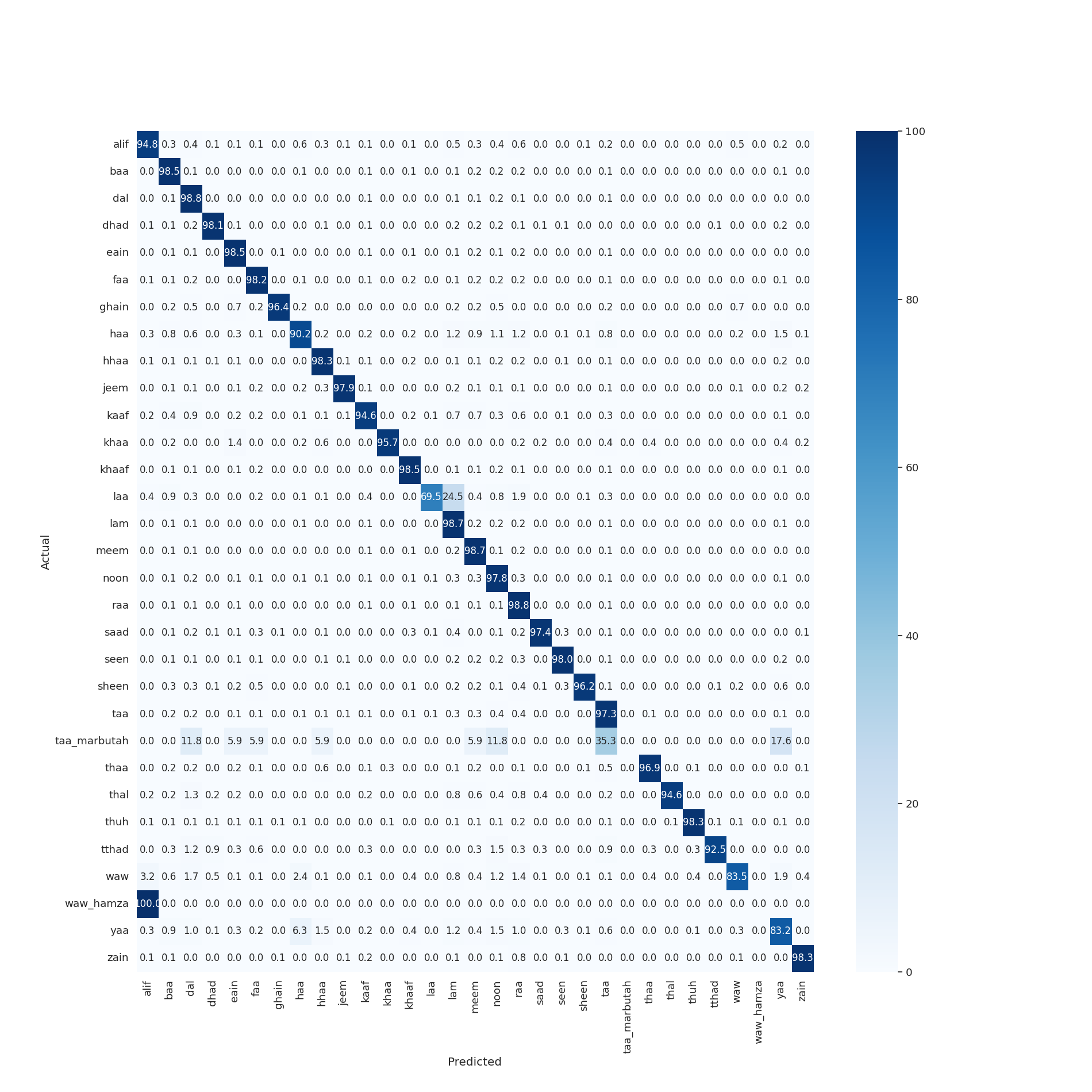}
\caption{Confusion Matrix for the test set in the rhyme classification task using AraPoemBERT.\label{fig11}}
\end{figure}  


In this task, all models have achieved similar results. CAMelBERT-CA scored the highest with an accuracy of 97.76\%, and AraPoemBERT achieved an accuracy of 97.73\%. Figure \ref{fig11} presents the confusion matrix for the validation set prediction results of AraPoemBERT which shows that the model can accurately identify the  rhymes if it is one of the original 28 Arabic letters. The remaining three rhymes (Laa, Taa Marbutah, and Waw Hamza) which are variants of the letters (Lam, Taa, and Alif), are where the model scores the lowest. This is due to multiple reasons. For instance, the number of samples for these rhymes is very small especially for "Taa Marbutah" and "Waw Hamza" rhymes. Also, the model wrongly classifies "Laa" as "Lam" for 24.5\% of the samples.

\section{Conclusion}\label{sec6}

In this study, we presented AraPoemBERT, a new BERT-based language model pretrained from scratch on Arabic poetry text. In addition, we employed the proposed model along with the other Arabic language model on five different NLP tasks related to Arabic poetry.  
The target tasks include classifying poets' genders, classifying poetry meters and sub-meters, sentiment analysis, and detecting verses' rhymes. The presented results illustrate the effectiveness of utilizing transformer-based models in various tasks related to Arabic poetry, and the significance of using a domain-specific language model such as AraPoemBERT that was exclusively pretrained on poetry text compared to language models pretrained on general text such as AraBERT and CAMeLBERT. The model has achieved state-of-the-art results and outperformed the other language models in most of the tasks. Also, we have explored three new NLP tasks in Arabic poetry that have not been explored in literature before: classifying poets' genders, classifying sub-meters, and detecting verses' rhymes. The results achieved in these tasks will serve as a benchmark for future work. Additionally, more NLP tasks related to Arabic poetry should be explored, such as authorship attribution, era classification, automating the process of poem text diacritization, and distinguishing between poems written in standard or spoken Arabic. The dataset and the language model introduced in this paper will serve as valuable resources for future work in different domains and fields such as linguistics, artificial intelligence, Arabic literature, language processing, and cultural studies.

\backmatter








\section*{Declarations}




\subsection{Conflict of interest}
The author declares no conflict of interest.

\subsection{Availability of data and materials}
AraPoemBERT model is publicly available on \url{https://huggingface.co/faisalq/bert-base-arapoembert}. AraPoems dataset is available on \url{ https://doi.org/10.7910/DVN/PJPWOY}

\subsection{Code availability}
The code along with the results is available on \url{https://github.com/FaisalQarah/araPoemBERT}.

\bibliography{sn-bibliography}


\begin{thebibliography}{39}
\ifx \bisbn   \undefined \def \bisbn  #1{ISBN #1}\fi
\ifx \binits  \undefined \def \binits#1{#1}\fi
\ifx \bauthor  \undefined \def \bauthor#1{#1}\fi
\ifx \batitle  \undefined \def \batitle#1{#1}\fi
\ifx \bjtitle  \undefined \def \bjtitle#1{#1}\fi
\ifx \bvolume  \undefined \def \bvolume#1{\textbf{#1}}\fi
\ifx \byear  \undefined \def \byear#1{#1}\fi
\ifx \bissue  \undefined \def \bissue#1{#1}\fi
\ifx \bfpage  \undefined \def \bfpage#1{#1}\fi
\ifx \blpage  \undefined \def \blpage #1{#1}\fi
\ifx \burl  \undefined \def \burl#1{\textsf{#1}}\fi
\ifx \doiurl  \undefined \def \doiurl#1{\url{https://doi.org/#1}}\fi
\ifx \betal  \undefined \def \betal{\textit{et al.}}\fi
\ifx \binstitute  \undefined \def \binstitute#1{#1}\fi
\ifx \binstitutionaled  \undefined \def \binstitutionaled#1{#1}\fi
\ifx \bctitle  \undefined \def \bctitle#1{#1}\fi
\ifx \beditor  \undefined \def \beditor#1{#1}\fi
\ifx \bpublisher  \undefined \def \bpublisher#1{#1}\fi
\ifx \bbtitle  \undefined \def \bbtitle#1{#1}\fi
\ifx \bedition  \undefined \def \bedition#1{#1}\fi
\ifx \bseriesno  \undefined \def \bseriesno#1{#1}\fi
\ifx \blocation  \undefined \def \blocation#1{#1}\fi
\ifx \bsertitle  \undefined \def \bsertitle#1{#1}\fi
\ifx \bsnm \undefined \def \bsnm#1{#1}\fi
\ifx \bsuffix \undefined \def \bsuffix#1{#1}\fi
\ifx \bparticle \undefined \def \bparticle#1{#1}\fi
\ifx \barticle \undefined \def \barticle#1{#1}\fi
\bibcommenthead
\ifx \bconfdate \undefined \def \bconfdate #1{#1}\fi
\ifx \botherref \undefined \def \botherref #1{#1}\fi
\ifx \url \undefined \def \url#1{\textsf{#1}}\fi
\ifx \bchapter \undefined \def \bchapter#1{#1}\fi
\ifx \bbook \undefined \def \bbook#1{#1}\fi
\ifx \bcomment \undefined \def \bcomment#1{#1}\fi
\ifx \oauthor \undefined \def \oauthor#1{#1}\fi
\ifx \citeauthoryear \undefined \def \citeauthoryear#1{#1}\fi
\ifx \endbibitem  \undefined \def \endbibitem {}\fi
\ifx \bconflocation  \undefined \def \bconflocation#1{#1}\fi
\ifx \arxivurl  \undefined \def \arxivurl#1{\textsf{#1}}\fi
\csname PreBibitemsHook\endcsname

\bibitem[\protect\citeauthoryear{Zwettler}{1978}]{zwettler1978oral}
\begin{bbook}
\bauthor{\bsnm{Zwettler}, \binits{M.}}:
\bbtitle{Oral Tradition of Classical Arabic Poetry: Its Character and Implications}.
\bpublisher{The Ohio State University Press},
\blocation{Columbus, Ohio}
(\byear{1978})
\end{bbook}
\endbibitem

\bibitem[\protect\citeauthoryear{Arberry}{1965}]{arberry1965arabic}
\begin{bbook}
\bauthor{\bsnm{Arberry}, \binits{A.J.}}:
\bbtitle{Arabic Poetry: a Primer for Students}.
\bpublisher{CUP Archive},
\blocation{Cambridge}
(\byear{1965})
\end{bbook}
\endbibitem

\bibitem[\protect\citeauthoryear{Scott}{2010}]{scott2010pegs}
\begin{botherref}
\oauthor{\bsnm{Scott}, \binits{H.}}:
Pegs, cords, and ghuls: Meter of classical arabic poetry.
PhD thesis
(2010)
\end{botherref}
\endbibitem

\bibitem[\protect\citeauthoryear{Atiq}{1987}]{atiq1987elm}
\begin{botherref}
\oauthor{\bsnm{Atiq}, \binits{A.}}:
Elm al-arud wal qafiah (in arabic).
Dar Alnahda, Beirut, Lebanon
(1987)
\end{botherref}
\endbibitem

\bibitem[\protect\citeauthoryear{Ahmed Abdel~Aziz and Juma Al~Kubaisi}{2018}]{ahmad2018causes}
\begin{botherref}
\oauthor{\bsnm{Ahmed Abdel~Aziz}, \binits{A.M.D.}},
\oauthor{\bsnm{Juma Al~Kubaisi}, \binits{M.S.}}:
Types of causes in the ancient arabic prosodic lesson.(in arabic).
Journal of Anbar University for Languages \& Literature/Magallat Gami'at Al-Anbar Li-Lugat Wa-al-Adabl
(26)
(2018)
\end{botherref}
\endbibitem

\bibitem[\protect\citeauthoryear{Sowayan et~al.}{2010}]{saad2010nabati}
\begin{bbook}
\bauthor{\bsnm{Sowayan}, \binits{S.A.}},
\bauthor{\bsnm{Al-Shaibani}, \binits{M.}},
\bauthor{\bsnm{Al-Zumar}, \binits{S.}}:
\bbtitle{Nabati Poetry The Taste of the People and The Authority of The Text (in Arabic)}.
\bpublisher{Abu Dhabi Authority for Culture and Heritage},
\blocation{Abu Dhabi}
(\byear{2010})
\end{bbook}
\endbibitem

\bibitem[\protect\citeauthoryear{Alabbas et~al.}{2014}]{alabbas2014basrah}
\begin{barticle}
\bauthor{\bsnm{Alabbas}, \binits{M.}},
\bauthor{\bsnm{Khalaf}, \binits{Z.A.}},
\bauthor{\bsnm{Khashan}, \binits{K.M.}}:
\batitle{Basrah: an automatic system to identify the meter of arabic poetry}.
\bjtitle{Natural Language Engineering}
\bvolume{20}(\bissue{1}),
\bfpage{131}--\blpage{149}
(\byear{2014})
\end{barticle}
\endbibitem

\bibitem[\protect\citeauthoryear{Abuata and Al-Omari}{2018}]{abuata2018rule}
\begin{barticle}
\bauthor{\bsnm{Abuata}, \binits{B.}},
\bauthor{\bsnm{Al-Omari}, \binits{A.}}:
\batitle{A rule-based algorithm for the detection of arud meter in classical arabic poetry}.
\bjtitle{International Arab Journal of Information Technology}
\bvolume{15}(\bissue{4}),
\bfpage{1}--\blpage{5}
(\byear{2018})
\end{barticle}
\endbibitem

\bibitem[\protect\citeauthoryear{Alnagdawi et~al.}{2013}]{alnagdawi2013finding}
\begin{barticle}
\bauthor{\bsnm{Alnagdawi}, \binits{M.A.}},
\bauthor{\bsnm{Rashideh}, \binits{H.}},
\bauthor{\bsnm{Aburumman}, \binits{F.}}:
\batitle{Finding arabic poem meter using context free grammar}.
\bjtitle{Journal of Communication and Computer Engineering}
\bvolume{3}(\bissue{1}),
\bfpage{52}--\blpage{59}
(\byear{2013})
\end{barticle}
\endbibitem

\bibitem[\protect\citeauthoryear{Berkani et~al.}{2020}]{berkani2020pattern}
\begin{bchapter}
\bauthor{\bsnm{Berkani}, \binits{A.}},
\bauthor{\bsnm{Holzer}, \binits{A.}},
\bauthor{\bsnm{Stoffel}, \binits{K.}}:
\bctitle{Pattern matching in meter detection of arabic classical poetry}.
In: \bbtitle{2020 IEEE/ACS 17th International Conference on Computer Systems and Applications (AICCSA)},
pp. \bfpage{1}--\blpage{8}
(\byear{2020}).
\bcomment{IEEE}
\end{bchapter}
\endbibitem

\bibitem[\protect\citeauthoryear{Yousef et~al.}{2019}]{yousef2019learning}
\begin{botherref}
\oauthor{\bsnm{Yousef}, \binits{W.A.}},
\oauthor{\bsnm{Ibrahime}, \binits{O.M.}},
\oauthor{\bsnm{Madbouly}, \binits{T.M.}},
\oauthor{\bsnm{Mahmoud}, \binits{M.A.}}:
Learning meters of arabic and english poems with recurrent neural networks: a step forward for language understanding and synthesis.
arXiv preprint arXiv:1905.05700
(2019)
\end{botherref}
\endbibitem

\bibitem[\protect\citeauthoryear{}{}]{apcd}
\begin{botherref}
Arabic Poem Comprehensive Dataset (APCD).
\url{https://hci-lab.github.io/LearningMetersPoems/}
\end{botherref}
\endbibitem

\bibitem[\protect\citeauthoryear{Al-Shaibani et~al.}{2020}]{alShaibani2020meter}
\begin{barticle}
\bauthor{\bsnm{Al-Shaibani}, \binits{M.S.}},
\bauthor{\bsnm{Alyafeai}, \binits{Z.}},
\bauthor{\bsnm{Ahmad}, \binits{I.}}:
\batitle{Meter classification of arabic poems using deep bidirectional recurrent neural networks}.
\bjtitle{Pattern Recognition Letters}
\bvolume{136},
\bfpage{1}--\blpage{7}
(\byear{2020})
\end{barticle}
\endbibitem

\bibitem[\protect\citeauthoryear{Abandah et~al.}{2022}]{abandah2022classifying}
\begin{barticle}
\bauthor{\bsnm{Abandah}, \binits{G.A.}},
\bauthor{\bsnm{Khedher}, \binits{M.Z.}},
\bauthor{\bsnm{Abdel-Majeed}, \binits{M.R.}},
\bauthor{\bsnm{Mansour}, \binits{H.M.}},
\bauthor{\bsnm{Hulliel}, \binits{S.F.}},
\bauthor{\bsnm{Bisharat}, \binits{L.M.}}:
\batitle{Classifying and diacritizing arabic poems using deep recurrent neural networks}.
\bjtitle{Journal of King Saud University-Computer and Information Sciences}
\bvolume{34}(\bissue{6}),
\bfpage{3775}--\blpage{3788}
(\byear{2022})
\end{barticle}
\endbibitem

\bibitem[\protect\citeauthoryear{Abboushi and Azzeh}{2023}]{abboushi2023toward}
\begin{botherref}
\oauthor{\bsnm{Abboushi}, \binits{O.}},
\oauthor{\bsnm{Azzeh}, \binits{M.}}:
Toward fluent arabic poem generation based on fine-tuning aragpt2 transformer.
Arabian Journal for Science and Engineering,
1--13
(2023)
\end{botherref}
\endbibitem

\bibitem[\protect\citeauthoryear{Antoun et~al.}{2020}]{antoun2020aragpt2}
\begin{botherref}
\oauthor{\bsnm{Antoun}, \binits{W.}},
\oauthor{\bsnm{Baly}, \binits{F.}},
\oauthor{\bsnm{Hajj}, \binits{H.}}:
Aragpt2: Pre-trained transformer for arabic language generation.
arXiv preprint arXiv:2012.15520
(2020)
\end{botherref}
\endbibitem

\bibitem[\protect\citeauthoryear{Mohammad}{2009}]{mohammad2009naive}
\begin{barticle}
\bauthor{\bsnm{Mohammad}, \binits{I.}}:
\batitle{Naive bayes for classical arabic poetry classification}.
\bjtitle{Al-Nahrain Journal of Science}
\bvolume{12}(\bissue{4}),
\bfpage{217}--\blpage{225}
(\byear{2009})
\end{barticle}
\endbibitem

\bibitem[\protect\citeauthoryear{Alsharif et~al.}{2013}]{alsharif2013emotion}
\begin{botherref}
\oauthor{\bsnm{Alsharif}, \binits{O.}},
\oauthor{\bsnm{Alshamaa}, \binits{D.}},
\oauthor{\bsnm{Ghneim}, \binits{N.}}:
Emotion classification in arabic poetry using machine learning.
International Journal of Computer Applications
\textbf{65}(16)
(2013)
\end{botherref}
\endbibitem

\bibitem[\protect\citeauthoryear{Ahmed et~al.}{2019}]{ahmed2019classification}
\begin{barticle}
\bauthor{\bsnm{Ahmed}, \binits{M.A.}},
\bauthor{\bsnm{Hasan}, \binits{R.A.}},
\bauthor{\bsnm{Ali}, \binits{A.H.}},
\bauthor{\bsnm{Mohammed}, \binits{M.A.}}:
\batitle{The classification of the modern arabic poetry using machine learning}.
\bjtitle{TELKOMNIKA (Telecommunication Computing Electronics and Control)}
\bvolume{17}(\bissue{5}),
\bfpage{2667}--\blpage{2674}
(\byear{2019})
\end{barticle}
\endbibitem

\bibitem[\protect\citeauthoryear{Shahriar et~al.}{2023}]{shahriar2023classification}
\begin{barticle}
\bauthor{\bsnm{Shahriar}, \binits{S.}},
\bauthor{\bsnm{Al~Roken}, \binits{N.}},
\bauthor{\bsnm{Zualkernan}, \binits{I.}}:
\batitle{Classification of arabic poetry emotions using deep learning}.
\bjtitle{Computers}
\bvolume{12}(\bissue{5}),
\bfpage{89}
(\byear{2023})
\end{barticle}
\endbibitem

\bibitem[\protect\citeauthoryear{Antoun et~al.}{2020}]{antoun2020arabert}
\begin{botherref}
\oauthor{\bsnm{Antoun}, \binits{W.}},
\oauthor{\bsnm{Baly}, \binits{F.}},
\oauthor{\bsnm{Hajj}, \binits{H.}}:
Arabert: Transformer-based model for arabic language understanding.
arXiv preprint arXiv:2003.00104
(2020)
\end{botherref}
\endbibitem

\bibitem[\protect\citeauthoryear{Vaswani et~al.}{2017}]{vaswani2017attention}
\begin{botherref}
\oauthor{\bsnm{Vaswani}, \binits{A.}},
\oauthor{\bsnm{Shazeer}, \binits{N.}},
\oauthor{\bsnm{Parmar}, \binits{N.}},
\oauthor{\bsnm{Uszkoreit}, \binits{J.}},
\oauthor{\bsnm{Jones}, \binits{L.}},
\oauthor{\bsnm{Gomez}, \binits{A.N.}},
\oauthor{\bsnm{Kaiser}, \binits{{\L}.}},
\oauthor{\bsnm{Polosukhin}, \binits{I.}}:
Attention is all you need.
Advances in neural information processing systems
\textbf{30}
(2017)
\end{botherref}
\endbibitem

\bibitem[\protect\citeauthoryear{El-Khair}{2016}]{1.5bwords}
\begin{botherref}
\oauthor{\bsnm{El-Khair}, \binits{I.A.}}:
1.5 billion words arabic corpus.
arXiv preprint arXiv:1611.04033
(2016)
\end{botherref}
\endbibitem

\bibitem[\protect\citeauthoryear{Zeroual et~al.}{2019}]{osian2019zeroual}
\begin{bchapter}
\bauthor{\bsnm{Zeroual}, \binits{I.}},
\bauthor{\bsnm{Goldhahn}, \binits{D.}},
\bauthor{\bsnm{Eckart}, \binits{T.}},
\bauthor{\bsnm{Lakhouaja}, \binits{A.}}:
\bctitle{Osian: Open source international arabic news corpus-preparation and integration into the clarin-infrastructure}.
In: \bbtitle{Proceedings of the Fourth Arabic Natural Language Processing Workshop},
pp. \bfpage{175}--\blpage{182}
(\byear{2019})
\end{bchapter}
\endbibitem

\bibitem[\protect\citeauthoryear{Abdelali et~al.}{2016}]{abdelali2016farasa}
\begin{bchapter}
\bauthor{\bsnm{Abdelali}, \binits{A.}},
\bauthor{\bsnm{Darwish}, \binits{K.}},
\bauthor{\bsnm{Durrani}, \binits{N.}},
\bauthor{\bsnm{Mubarak}, \binits{H.}}:
\bctitle{Farasa: A fast and furious segmenter for arabic}.
In: \bbtitle{Proceedings of the 2016 Conference of the North American Chapter of the Association for Computational Linguistics: Demonstrations},
pp. \bfpage{11}--\blpage{16}
(\byear{2016})
\end{bchapter}
\endbibitem

\bibitem[\protect\citeauthoryear{Chowdhury et~al.}{2020}]{chowdhury2020qarib}
\begin{bchapter}
\bauthor{\bsnm{Chowdhury}, \binits{S.A.}},
\bauthor{\bsnm{Abdelali}, \binits{A.}},
\bauthor{\bsnm{Darwish}, \binits{K.}},
\bauthor{\bsnm{Soon-Gyo}, \binits{J.}},
\bauthor{\bsnm{Salminen}, \binits{J.}},
\bauthor{\bsnm{Jansen}, \binits{B.J.}}:
\bctitle{Improving arabic text categorization using transformer training diversification}.
In: \bbtitle{Proceedings of the Fifth Arabic Natural Language Processing Workshop},
pp. \bfpage{226}--\blpage{236}
(\byear{2020})
\end{bchapter}
\endbibitem

\bibitem[\protect\citeauthoryear{Abdul-Mageed et~al.}{2020}]{abdul2020arbert}
\begin{botherref}
\oauthor{\bsnm{Abdul-Mageed}, \binits{M.}},
\oauthor{\bsnm{Elmadany}, \binits{A.}},
\oauthor{\bsnm{Nagoudi}, \binits{E.M.B.}}:
Arbert \& marbert: deep bidirectional transformers for arabic.
arXiv preprint arXiv:2101.01785
(2020)
\end{botherref}
\endbibitem

\bibitem[\protect\citeauthoryear{Su{\'a}rez et~al.}{2019}]{oscar2019asynchronous}
\begin{bchapter}
\bauthor{\bsnm{Su{\'a}rez}, \binits{P.J.O.}},
\bauthor{\bsnm{Sagot}, \binits{B.}},
\bauthor{\bsnm{Romary}, \binits{L.}}:
\bctitle{Asynchronous pipeline for processing huge corpora on medium to low resource infrastructures}.
In: \bbtitle{7th Workshop on the Challenges in the Management of Large Corpora (CMLC-7)}
(\byear{2019}).
\bcomment{Leibniz-Institut f{\"u}r Deutsche Sprache}
\end{bchapter}
\endbibitem

\bibitem[\protect\citeauthoryear{Inoue et~al.}{2021}]{inoue2021camelbert}
\begin{botherref}
\oauthor{\bsnm{Inoue}, \binits{G.}},
\oauthor{\bsnm{Alhafni}, \binits{B.}},
\oauthor{\bsnm{Baimukan}, \binits{N.}},
\oauthor{\bsnm{Bouamor}, \binits{H.}},
\oauthor{\bsnm{Habash}, \binits{N.}}:
The interplay of variant, size, and task type in arabic pre-trained language models.
arXiv preprint arXiv:2103.06678
(2021)
\end{botherref}
\endbibitem

\bibitem[\protect\citeauthoryear{Alammary}{2022}]{alammary2022bert}
\begin{barticle}
\bauthor{\bsnm{Alammary}, \binits{A.S.}}:
\batitle{Bert models for arabic text classification: a systematic review}.
\bjtitle{Applied Sciences}
\bvolume{12}(\bissue{11}),
\bfpage{5720}
(\byear{2022})
\end{barticle}
\endbibitem

\bibitem[\protect\citeauthoryear{Devlin et~al.}{2018}]{devlin2018bert}
\begin{botherref}
\oauthor{\bsnm{Devlin}, \binits{J.}},
\oauthor{\bsnm{Chang}, \binits{M.-W.}},
\oauthor{\bsnm{Lee}, \binits{K.}},
\oauthor{\bsnm{Toutanova}, \binits{K.}}:
Bert: Pre-training of deep bidirectional transformers for language understanding.
arXiv preprint arXiv:1810.04805
(2018)
\end{botherref}
\endbibitem

\bibitem[\protect\citeauthoryear{Wu et~al.}{2016}]{wordpiece}
\begin{botherref}
\oauthor{\bsnm{Wu}, \binits{Y.}},
\oauthor{\bsnm{Schuster}, \binits{M.}},
\oauthor{\bsnm{Chen}, \binits{Z.}},
\oauthor{\bsnm{Le}, \binits{Q.V.}},
\oauthor{\bsnm{Norouzi}, \binits{M.}},
\oauthor{\bsnm{Macherey}, \binits{W.}},
\oauthor{\bsnm{Krikun}, \binits{M.}},
\oauthor{\bsnm{Cao}, \binits{Y.}},
\oauthor{\bsnm{Gao}, \binits{Q.}},
\oauthor{\bsnm{Macherey}, \binits{K.}}, et al.:
Google's neural machine translation system: Bridging the gap between human and machine translation.
arXiv preprint arXiv:1609.08144
(2016)
\end{botherref}
\endbibitem

\bibitem[\protect\citeauthoryear{}{}]{w1}
\begin{botherref}
Almausua.
\url{https://poetry.dctabudhabi.ae/}.
Accessed: 2023-01-10
\end{botherref}
\endbibitem

\bibitem[\protect\citeauthoryear{}{}]{w2}
\begin{botherref}
Aldiwan.
\url{https://www.aldiwan.net/}.
Accessed: 2023-01-19
\end{botherref}
\endbibitem

\bibitem[\protect\citeauthoryear{Wolf et~al.}{2020}]{wolf2020transformers}
\begin{bchapter}
\bauthor{\bsnm{Wolf}, \binits{T.}},
\bauthor{\bsnm{Debut}, \binits{L.}},
\bauthor{\bsnm{Sanh}, \binits{V.}},
\bauthor{\bsnm{Chaumond}, \binits{J.}},
\bauthor{\bsnm{Delangue}, \binits{C.}},
\bauthor{\bsnm{Moi}, \binits{A.}},
\bauthor{\bsnm{Cistac}, \binits{P.}},
\bauthor{\bsnm{Rault}, \binits{T.}},
\bauthor{\bsnm{Louf}, \binits{R.}},
\bauthor{\bsnm{Funtowicz}, \binits{M.}}, \betal:
\bctitle{Transformers: State-of-the-art natural language processing}.
In: \bbtitle{Proceedings of the 2020 Conference on Empirical Methods in Natural Language Processing: System Demonstrations},
pp. \bfpage{38}--\blpage{45}
(\byear{2020})
\end{bchapter}
\endbibitem

\bibitem[\protect\citeauthoryear{Zerrouki}{2010}]{zerrouki2010pyarabic}
\begin{botherref}
\oauthor{\bsnm{Zerrouki}, \binits{T.}}:
Pyarabic, an Arabic language library for Python.
Pyarabic
(2010)
\end{botherref}
\endbibitem

\bibitem[\protect\citeauthoryear{Liu et~al.}{2019}]{liu2019roberta}
\begin{botherref}
\oauthor{\bsnm{Liu}, \binits{Y.}},
\oauthor{\bsnm{Ott}, \binits{M.}},
\oauthor{\bsnm{Goyal}, \binits{N.}},
\oauthor{\bsnm{Du}, \binits{J.}},
\oauthor{\bsnm{Joshi}, \binits{M.}},
\oauthor{\bsnm{Chen}, \binits{D.}},
\oauthor{\bsnm{Levy}, \binits{O.}},
\oauthor{\bsnm{Lewis}, \binits{M.}},
\oauthor{\bsnm{Zettlemoyer}, \binits{L.}},
\oauthor{\bsnm{Stoyanov}, \binits{V.}}:
Roberta: A robustly optimized bert pretraining approach.
arXiv preprint arXiv:1907.11692
(2019)
\end{botherref}
\endbibitem

\bibitem[\protect\citeauthoryear{Loshchilov and Hutter}{2017}]{loshchilov2017adamw}
\begin{botherref}
\oauthor{\bsnm{Loshchilov}, \binits{I.}},
\oauthor{\bsnm{Hutter}, \binits{F.}}:
Decoupled weight decay regularization.
arXiv preprint arXiv:1711.05101
(2017)
\end{botherref}
\endbibitem

\bibitem[\protect\citeauthoryear{Zhang et~al.}{2020}]{zhang2020you}
\begin{botherref}
\oauthor{\bsnm{Zhang}, \binits{Y.}},
\oauthor{\bsnm{Warstadt}, \binits{A.}},
\oauthor{\bsnm{Li}, \binits{H.-S.}},
\oauthor{\bsnm{Bowman}, \binits{S.R.}}:
When do you need billions of words of pretraining data?
arXiv preprint arXiv:2011.04946
(2020)
\end{botherref}
\endbibitem

\end{thebibliography}

\end{document}